\documentclass[runningheads]{llncs}

\usepackage{eccv}

\usepackage{eccvabbrv}

\usepackage{graphicx}
\usepackage{booktabs}
\usepackage{xcolor}
\usepackage{arydshln}
\usepackage{amsmath,amssymb}
\usepackage{xspace}
\usepackage{graphicx}
\usepackage{subcaption}
\usepackage{dashrule}
\usepackage{multirow}
\usepackage{colortbl}
\usepackage{booktabs}
\usepackage{cancel}
\usepackage{microtype}
\usepackage{comment}

\usepackage{algorithm}
\usepackage{algorithmic}

\def\eg{\textit{e.g.}} 
\def\ie{\textit{i.e.}}

\newcommand{\ec}[2][EC: ]{{\color{black}#1#2}}
\newcommand{\fb}[2][FB: ]{{\color{black}#1#2}}

\definecolor{lightblue}{RGB}{173, 216, 230}
\definecolor{lightgreen}{RGB}{173, 230, 216}
\definecolor{lightred}{RGB}{230, 173, 216}
\definecolor{lightyellow}{RGB}{230, 230, 173}

\definecolor{blueish}{RGB}{232, 250, 250}
\definecolor{grayish}{RGB}{238, 238, 244}
\definecolor{yellyish}{RGB}{250, 245, 225}

\definecolor{mocha-light}{RGB}{217, 179, 130}
\definecolor{mocha-mid}{RGB}{160, 81, 45}
\definecolor{mocha-dark}{RGB}{75, 56, 50}

\newcommand{\pms}[1]{\color{mocha-dark!70}\rm\fontsize{5}{10}\ensuremath{\pm #1}}

\newcommand{\rev}[1]{\textcolor{black}{#1}}
\newcommand{\revv}[1]{\textcolor{black}{#1}}

\usepackage[accsupp]{axessibility}  %

\usepackage{hyperref}

\usepackage{orcidlink}

\begin{document}

\title{MOCHA: Multi-modal Objects-aware Cross-arcHitecture Alignment} 

\titlerunning{MOCHA: Multi-modal Objects-aware Cross-arcHitecture Alignment}

\author{Elena Camuffo\inst{1,2}\orcidlink{0000-0002-8351-4650} \and
Francesco Barbato\inst{1,2}\orcidlink{0000-0001-9893-5813} \and
Mete Ozay\inst{1}\orcidlink{0000-0002-7189-7260} \and
Simone Milani\inst{2}\orcidlink{0000-0001-8266-5839} \and
Umberto Michieli\inst{1}\orcidlink{0000-0003-2666-4342}}

\authorrunning{E.~Camuffo et al.}

\institute{Samsung R\&D Institute UK, United Kingdom \and
University of Padova, Italy\\
\smallskip
\url{https://github.com/SamsungLabs/MOCHA}} 

\maketitle

\begin{abstract}
Personalized object detection aims to adapt a general-purpose detector to recognize user-specific instances from only a few examples.
Lightweight models often struggle in this setting due to their weak semantic priors, while large vision-language models (VLMs) offer strong object-level understanding but are too computationally demanding for real-time or on-device applications.
We introduce MOCHA (Multi-modal Objects-aware Cross-arcHitecture Alignment), a distillation framework that transfers multimodal region-level knowledge from a frozen VLM teacher into a lightweight vision-only detector. 
MOCHA extracts fused visual and textual teacher's embeddings and uses them to guide student training through a dual-objective loss that enforces accurate local alignment and global relational consistency across regions.
This process enables efficient transfer of semantics without the need for teacher modifications or textual input at inference.
MOCHA consistently outperforms prior baselines across four personalized detection benchmarks under strict few-shot regimes, yielding a +10.1 average improvement, with minimal inference cost.

\end{abstract}

\section{Introduction}

\begin{figure}[t]
    \centering
    \includegraphics[width=0.9\linewidth]{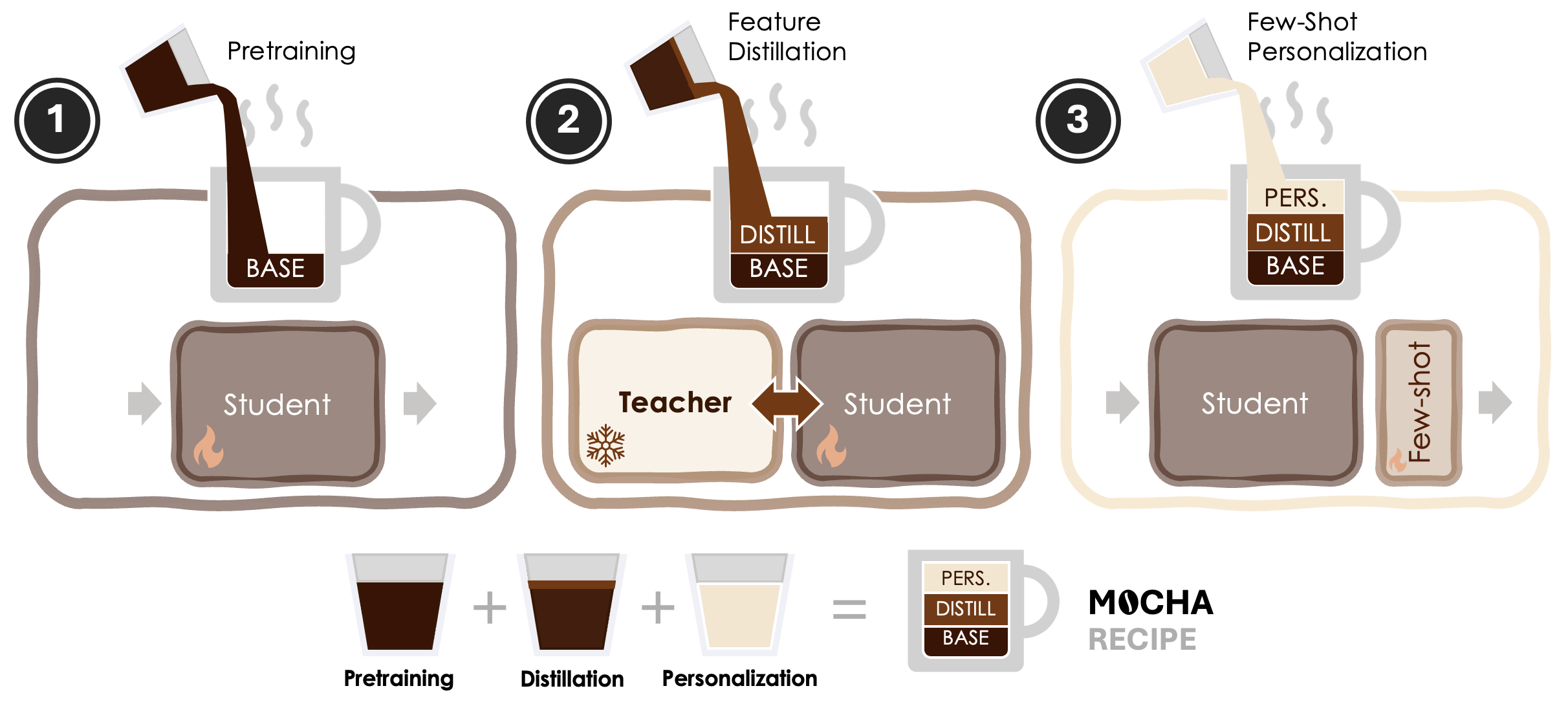}
    \caption{\textbf{MOCHA recipe.} (1) Pretraining student model. (2) Knowledge distillation on rich joint visual and textual features from a frozen teacher. (3) Few-shot personalization with frozen student and prototypical learner.}
    \label{fig:graph-abs}
\end{figure}

Recent advances in vision-language models (VLMs) such as CLIP~\cite{radford2021learning}, Flamingo~\cite{alayrac2022flamingo}, and LLaVa~\cite{liu2023LLaVa} have demonstrated remarkable zero-shot and open-vocabulary %
capabilities, owing to their rich %
representations and scale. These models, however, are typically large and computationally intensive, limiting their applicability in real-time or resource-constrained scenarios such as mobile devices (\eg, smartphones or robots).
In contrast, lightweight architectures such as YOLO~\cite{yolov8_ultralytics,yolo11_ultralytics} offer fast and memory-efficient object detection, at the cost of degraded performance in low-data regimes and vulnerability to neural collapse~\cite{Kothapalli2022NeuralCA,papyan2020neural}, where features become indistinguishably aligned and lose their discriminative power.
In this work, we aim to bridge this gap by proposing {MOCHA}, a knowledge distillation approach that transfers object-centric multimodal embeddings from a vision-language teacher %
into a compact vision-only student detector. 
MOCHA {builds on the observation} that semantically related concepts tend to exhibit similar structures in the embedding space across modalities under a well-generalized backbone \cite{huh2024platonicrepresentationhypothesis,girdhar2023imagebind,camuffo2023learning}.
By aligning and regularizing the student's feature space, it achieves improved generalization in few-shot detection settings.
{Our approach consists of three stages (Fig.~\ref{fig:graph-abs}): (1) pretraining a student detector, (2) distilling rich joint visual-textual features from a frozen teacher, and (3) performing few-shot personalization with a frozen student and a prototype-based classifier \cite{snell2017prototypical}.}

Our primary contribution lies in the distillation stage, where we introduce: 
(i) extraction of joint visual-language features from a frozen teacher, using both image regions and class labels to provide strong object-level supervision;
(ii) a translation module that maps the student's visual features into the teacher's multimodal space; and
(iii) a distillation objective that combines local feature alignment with a relational constraint, helping the student match the teacher's semantics and get relationships between object regions.

\revv{Unlike prior cross-modal or cross-architecture distillation approaches (\eg, AuXFT~\cite{barbato2024crossarchitectureauxiliaryfeaturespace}, OFA~\cite{hao2023oneforall}, GLIP~\cite{li2022glip}), MOCHA targets object-level few-shot personalized detection by aligning lightweight detector features with compact multimodal embeddings from a frozen VLM teacher.
In particular, unlike AuXFT's dense visual feature transfer between related visual architectures, MOCHA bridges both a modality gap (vision-only vs. vision-language) and a structural mismatch (multi-scale detector features vs. object-centric VLM representations) through multimodal fusion and relational alignment.}

We validate MOCHA on personal object detection datasets, where the goal is to adapt a general detector to user-specific personal classes (\eg, distinguishing a generic \textit{dog} from a particular \textit{user's dog}). MOCHA achieves significant improvements over state-of-the-art methods while retaining minimal inference cost, making it practical for resource-constrained deployment.

\section{Related Work}

\paragraph{Cross-Architecture Knowledge Distillation.} 

Traditional knowledge distillation methods~\cite{fitnets,zhang2020task,passalis2020heterogeneous,luo2019knowledge,10465265,9834142,10551493,10849981,li2022glip,ma2023skdf,ni2023drkd} typically assume that teacher and student networks share similar architectures. The adoption of large transformer-based models~\cite{vaswani2017attention,dosovitskiy2020image,devlin2018bert,liu2021swin} has raised new challenges in transferring knowledge into efficient CNNs, which offer better trade-offs for deployment. Recent works~\cite{liu2022cross,hao2023oneforall} address this by projecting multi-scale features from teacher and student into a shared space for supervision. 
These methods primarily aim to improve performance on the same task shared by teacher and student. In contrast, our approach aligns more closely with AuXFT~\cite{barbato2024crossarchitectureauxiliaryfeaturespace}, which leverages teacher features to guide a student on a related auxiliary task without degrading performance on the original one. Specifically, AuXFT addresses few-shot instance-level personalized object detection by injecting high-level teacher features into a lightweight detector to support prototype-based classification.
\ec[]{
To the best of our knowledge, AuXFT is the only prior work exploring a similar setup.}
Our method differs in three ways: (i) we leverage multimodal supervision from both visual and textual signals, unlike purely visual clues used in prior art; (ii) we distill compact region-level embeddings rather than dense intermediate activations, improving efficiency; and (iii) we explicitly regularize the embedding space to encourage geometric and relational alignment between individual instances.

\paragraph{Cross-Modal Adaptation.}
Several cross-modal learning strategies have been proposed to exploit complementary information across modalities \cite{xia2023cmda,barbato2024continual} or to enforce consistency between predictions from different modalities~\cite{jaritz2022cross,jaritz2020xmuda}.
\revv{In our work, the VLM is used only offline as a teacher: after distillation, MOCHA performs personalization and detection with a compact vision-only detector, without text prompts or VLM inference at test time.}

\paragraph{Personalized Scene Understanding.}
Personalization was first explored in Natural Language Processing (NLP)~\cite{liu2023LLaVa,vaswani2017attention,devlin2018bert}. %
Later, it was extended to computer vision, with early applications in personalized semantic segmentation~\cite{zhang2021personalized} and object detection~\cite{michieli2024object}. 
The advent of foundation models such as CLIP~\cite{radford2021learning} and SAM~\cite{kirillov2023segment} has shifted personalization strategies toward prompt-based control of generalist architectures~\cite{10507865}. 
 Building on this trend, several methods leverage strong visual or multimodal representations to guide instance-aware predictions through textual or visual prompts, including open-vocabulary detectors like ViLD~\cite{vild2021} and prompt-driven approaches such as SwissDINO~\cite{paramonov2024swiss}, PerSAM~\cite{zhang2024personalize}, Matcher~\cite{liu2023matcher}, and SegGPT~\cite{wang2023seggpt}.
\ec[]{While these approaches allow for zero-shot recognition, they are computationally demanding and do not directly address the few-shot instance-level recognition scenario considered here.} %
Our work instead distills rich semantic features into a compact detector during server-side training, avoiding reliance on \fb[]{textual prompts at test time (that would add significant computational costs)} or online adaptation, and enabling efficient few-shot personalization for resource-constrained deployment.

\begin{figure*}[t]
    \centering
    \includegraphics[width=\linewidth]{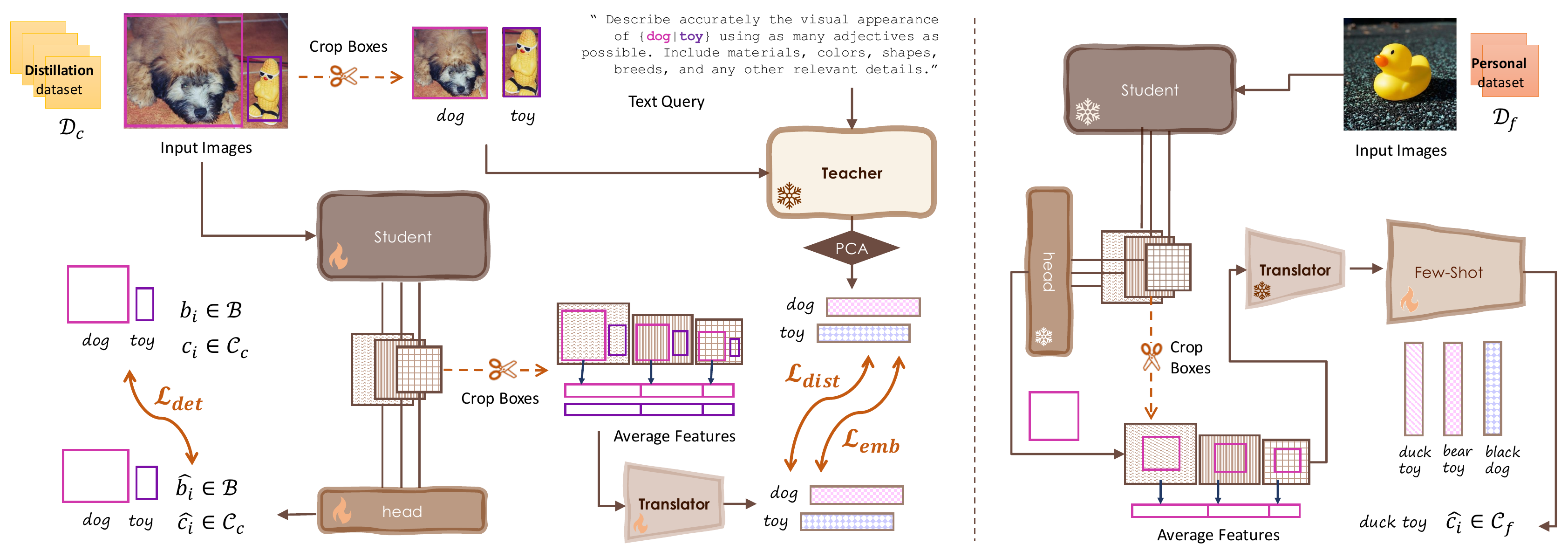}
    \begin{minipage}[b]{0.65\linewidth}
        \centering
        \subcaption{Feature distillation stage of the student network from the frozen teacher. %
        }\label{fig:pretraining}
    \end{minipage}%
    \hfill
    \begin{minipage}[b]{0.35\linewidth}
        \centering
        \subcaption{Personalization stage.}\label{fig:personalization}
    \end{minipage}
    \caption{\textbf{MOCHA system.} {(a) \textit{Feature distillation:} Student detector is trained on dataset $\mathcal{D}_c$ aligning multiscale region-level features to PCA-pruned multimodal embeddings from a frozen vision-language teacher via the translation module $\mathrm{t}_S(\cdot)$.  %
    (b) \textit{Personalization:} Student backbone and Translator are frozen and used to compute semantic prototypical features from a personal dataset $\mathcal{D}_f$. These are then used to train a prototype-based few-shot learner $\mathrm{p}(\cdot)$ for user-specific object classification.}
    }
    \label{fig:pipeline}
\end{figure*}

\section{Methodology}

{MOCHA} (\underline{M}ulti-modal \underline{O}bjects-aware \underline{C}ross-arc\underline{H}itecture \underline{A}lignment) distills multimodal semantics from a vision-language foundation model into a lightweight student detector by aligning region-level embeddings and regularizing the student feature space.
MOCHA enhances feature separability and generalization for few-shot personalized detection. 
Our method is architecture-agnostic and combines multimodal supervision, feature translation, and a dual loss enforcing both local and relational alignment.

\subsection{Problem Setup}
MOCHA 
consists of three stages, as described next.

\paragraph{Base Pretraining.}
We begin with a standard detection model $\mathrm{m}_{S}(\cdot) = \mathrm{l}_{S} \circ \mathrm{g}_{S}$, where $\mathrm{l}_{S}(\cdot)$ is the detection head, $\mathrm{g}_{S}(\cdot)$ is the student backbone, and $\circ$ is the composition operator. We train it on a pretraining detection dataset using standard objectives $\mathcal{L}_{det}$ \cite{yolov8_ultralytics}. This stage results in a strong generic detector that identifies broad object categories but lacks personalized semantics or fine-grained embedding separation and serves as initialization for the student's weights.

\paragraph{Feature Distillation.}
After initialization, we perform feature distillation over an object detection dataset $\mathcal{D}_c$, using a frozen large vision-language model as the teacher (Fig.~\ref{fig:pretraining}). 
Each image $X \in \mathcal{D}_c$ in the dataset is labeled by a set of tuples $\{(b_i, c_i)\}_{i=1}^n \subset \mathcal{B} \times \mathcal{C}_c$, where $c_i \in \mathcal{C}_c$ are class labels and $b_i \in \mathcal{B} \subset \mathbb{R}^4$ are %
box coordinates identifying $n$ image regions $X_i$.
The teacher %
is assumed to %
comprise a visual encoder $\mathrm{g}_{T}(\cdot)$, and a large language model $\mathrm{f}_{T}(\cdot)$, enabling the joint processing of images and text.
Note that, in general, the student and teacher features may not be aligned. For this reason, we use a feature translation module $\mathrm{t}_{S}(\cdot): \mathbb{R}^{d_s} \mapsto \mathbb{R}^{d_t}$ (where $d_s$ and $d_t$ are the dimensions of the student and teacher features, respectively), to allow knowledge distillation from the student's to the teacher's space.
\revv{Importantly, teacher embeddings are precomputed and cached, avoiding VLM inference during training iterations and deployment.}

\paragraph{Few-Shot Personalization.}
After feature distillation, the student backbone is frozen and used to extract multiscale region-level features $\{F_j\}_j = \mathrm{g}_{S}(X)$ from a personal detection dataset $\mathcal{D}_f$, labeled with bounding boxes $\mathcal{B}$ and class labels $\mathcal{C}_f$. 
The features are translated into the shared embedding space via the frozen translation module (Fig.~\ref{fig:personalization}) and fed to a prototype-based few-shot learner $\mathrm{p}(\cdot)$, like nearest class mean \cite{snell2017prototypical}, which is trained and tested as in \cite{barbato2024crossarchitectureauxiliaryfeaturespace}.
More in detail, during personalization, a user provides a few labeled samples of personal instances of objects to the system (\eg, 1-5 samples).
Later, during inference, the system automatically overrides the coarse predictions with the personal labels.
\revv{This strategy enables the student to adapt to novel visual concepts while retaining its compact architecture.
The prototype-based classifier computes class-specific prototypes from distilled embeddings without teacher inference, prompt engineering, or detector fine-tuning, enabling low-overhead personalization.}

\subsection{Multimodal Supervision Extraction}
Without loss of generality, in our experiments, we instantiate the teacher with a LLaVa~\cite{liu2023LLaVa} model, which includes a shared projection matrix $W_T$, in addition to $\mathrm{g}_T(\cdot)$ and $\mathrm{f}_T(\cdot)$.
Each cropped image region $X_i$ is processed by the teacher's visual encoder to obtain a visual embedding $Z_{V,i}$, %
projected into a joint text-vision representation space as $H_{V,i} = W_T Z_{V,i}$. Simultaneously, the class label $c_i$ is embedded as a textual query $Z_{Q,i}$ and %
projected via $H_{Q,i} = W_T Z_{Q,i}$.
The pair $(H_{V,i}, H_{Q,i})$ is passed to the frozen language model $\mathrm{f}_T$, which computes a sequence of tokens describing the interaction between the region and the class.  
We average the output along the temporal dimension to obtain a single semantic embedding $h_i \in \mathbb{R}^{d_h}$ representing the fused visual-linguistic description of the region $X_i$.

\paragraph{Dimensionality Reduction.}
While the teacher's multimodal feature $h_i$ captures rich semantics, it does not necessarily retain appearance-level cues. 
To combine both semantic and visual signals, we concatenate the multimodal representation with the $d_z$-dimensional class token $z_{V,i} \in Z_{V,i} \subset \mathbb{R}^{d_z}$ 
of its corresponding visual embedding:
\begin{equation}
    u_i = \mathrm{concat}(\gamma \ z_{V,i}, h_i) \in \mathbb{R}^d, \ \ \  d=d_z+d_h.
\end{equation}

with $\gamma = ||h_i||$ as a rescaling factor that normalizes visual features to textual ones.
\rev{
In practice, $h_i$ is obtained by mean-pooling the hidden states of the final decoder layer of the VLM, while $z_{V,i}$ corresponds to the $\ell_2$-normalized class token of the visual encoder. 
The scaling factor $\gamma$ balances the norms of visual and textual features, preventing the higher-magnitude modality from dominating the fused representation.
}
This combined vector represents both the raw visual content of the object and its semantic interpretation conditioned on the label. 
However, the resulting representation is high-dimensional and could include redundant information. To reduce dimensionality and improve efficiency, we apply Principal Component Analysis (PCA) to $u_i$, offline on the same dataset used for distillation, and we take the first $d_t$ elements:
\begin{equation}
    \hat{u}'_i = \mathrm{PCA}(u_i) \in \mathbb{R}^{d_t} \;.
\end{equation}
Finally, to equalize the contribution of each channel in the PCA representation, we estimate the standard deviation of each channel activation $\sigma_c$ in the distillation dataset, and use it to rescale the channels as:
\begin{equation}
    u'_i[c] = \hat{u}'_i[c] / \sigma_c \;\;\; \text{for } c\!=\!1, \dots, d.
\end{equation}
Remarkably, the deviations closely follow a hyperbolic shape, allowing us to estimate their value based only on the channel index (detailed discussion in the \ec[]{Appendix}).

\subsection{Student Detector and Feature Aggregation}

To perform object-level supervision, we apply the same cropping strategy used for the teacher: for each ground truth bounding box $b_i$, we first resize the feature maps to a common size to account for resolution differences, ensuring consistent spatial alignment. Then, we extract the corresponding region-aligned features $F_{A,j,i}$ from each resolution level $j$, and apply spatial average pooling to obtain fixed-size descriptors $f_{A,j,i} \in \mathbb{R}^{d_s}$.

The final aggregated feature for region $i$ is formed by concatenating the pooled representations at each level:
\begin{equation}
f_{A,i} = \mathrm{concat}_j({f}_{A,j,i}) \in \mathbb{R}^{d_s}. 
\end{equation} 
We denote with $F_A$ the dense equivalent of these features, that is, the concatenation without region pooling.

\paragraph{Feature Translation Module.}

The translation module $\mathrm{t}_S$ is modelled after a transformer encoder block and consists of a channel-wise multi-head self-attention block followed by a lightweight multi-layer perceptron (MLP).  
The attention mechanism enhances expressiveness by modelling inter-channel dependencies, while the MLP adapts the representation to the required dimensionality:
\begin{equation}
f'_{A,i} = \mathrm{t}_S(f_{A,i}),
\end{equation}
where $f'_{A,i} \in \mathbb{R}^{d_t}$ is the translated student feature with the same dimension as the PCA-compressed teacher target $u'_i$.

\rev{
Dimensionality matching alone is insufficient due to the heterogeneity between detector and VLM latent spaces. 
The translation module therefore performs a semantic mapping rather than a simple projection, learning to bridge the structural gap between region-level CNN features and multimodal transformer representations.
}
In our implementation (Fig.~\ref{fig:pretraining}), the translation module is trained jointly with the student detector, allowing the feature alignment to evolve progressively during distillation. 

\subsection{Training Objectives and Strategy}

To align the translated student features with the teacher targets, we define two complementary objectives: a pointwise distillation loss and a relational embedding loss.

\paragraph{Distillation Loss.}
The primary objective enforces direct alignment between translated student features $f'_{A,i}$ and their corresponding teacher vector $u'_i$.  
We define the distillation loss as the average of $\ell_1$ and $\ell_2$ distances, as in \cite{barbato2024crossarchitectureauxiliaryfeaturespace}:
\begin{equation}
\mathcal{L}_{\mathrm{dist}} = \frac{1}{n} \sum_{i=1}^n \left( \|f'_{A,i} - u'_i\|_1 + \|f'_{A,i} - u'_i\|_2 \right).
\end{equation}
This formulation captures both robustness to outliers (via the $\ell_1$ term) and fine-grained magnitude alignment (via the $\ell_2$ term), promoting stable and discriminative alignment.

\paragraph{Embedding Loss.}
Beyond individual alignment, we encourage the student to preserve the global geometry of the teacher's embedding space.  
To this end, we define a relational embedding loss based on pairwise Euclidean distances between all student and teacher features.
Given the translated student features $\{f'_{A,i}\}_i$ and the corresponding teacher targets $\{u'_i\}_i$, \ec[]{we compute pairwise distance matrices $D_{f\!f}, D_{uu}$ using normalized Euclidean distances. We remove self-distances by discarding diagonal entries and convert into probability distributions using softmax}: 
\begin{equation}
P_{f\!f} = \mathrm{softmax}\left(-D_{f\!f}\right), \ \  P_{uu} = \mathrm{softmax}\left(-D_{uu}\right).
\end{equation}
The embedding loss is then defined as the cross-entropy between the two distributions:
\begin{equation}\label{eq:embloss}
\mathcal{L}_{\mathrm{emb}} = -\frac{1}{n} \sum_{i=1}^n \sum_{j \neq i} P_{uu}[i,j] \log P_{f\!f}[i,j].
\end{equation}

This term ensures that the student preserves not only semantic content but also the relative structure of the embedding space, reflecting inter-class and intra-class relationships encoded by the teacher.

\rev{
Although the relational loss involves pairwise interactions and is theoretically $\mathcal{O}(N^2)$, in practice it is computed over a small set of ground-truth object regions per image, making the overhead negligible compared to standard convolutional operations (on OpenImages, we observe on average $N \approx 8$ regions per image). 
}

\paragraph{Final Objective.}

The final objective combines the detection loss with the distillation and embedding terms:
\begin{equation}
\mathcal{L} = \mathcal{L}_{\mathrm{det}} + \lambda_{\mathrm{dist}} \mathcal{L}_{\mathrm{dist}} + \lambda_{\mathrm{emb}} \mathcal{L}_{\mathrm{emb}},
\end{equation}
where the $\lambda$ coefficients regulate the contribution of each auxiliary term.
\ec[]{This combined objective ensures that the student learns both task-relevant representations and the semantic structure transferred from the teacher.}

\section{Experiments}
{We evaluate MOCHA in the context of few-shot personalized object detection, assessing its ability to transfer multimodal knowledge from a large VLM teacher into lightweight student detectors. Our experiments cover four personal datasets and analyze performance from three complementary perspectives: general behavior and design insights, comparison with state-of-the-art methods, and detailed ablation studies.
Results are reported after introducing the experimental setup and presenting preliminary analyses.}

\subsection{Experimental Setup} 
\paragraph{Datasets.} For pretraining, we adopt COCO~\cite{Lin2014MicrosoftCC} and OpenImagesV7 \cite{kuznetsova2020open}. Also, we consider starting from AuXFT pretrained weights (which in turn start from COCO and fine-tune on OpenImages).
Distillation is performed on OpenImages.
Considered personalization datasets are: PerSeg \cite{zhang2024personalize}, POD \cite{barbato2024crossarchitectureauxiliaryfeaturespace}, CORe50 \cite{lomonaco2017CORe50}, and iCubWorld \cite{Fanello2013CVPRws}. 

\paragraph{Models.}
Our approach uses YOLOv8n ($\approx3.2$M params) as the student model and LLaVa-1.5-7B ($\approx7.2$B params) as the teacher, which includes CLIP with ViT-B/32 as the visual encoder and LLaMA 7B as the text encoder.
For comparison, we also include: (i) the text encoder from LLaVa, \ie, LLaMA 7B ($\approx7$B params) (ii) the visual encoder from LLaVa, \ie, CLIP ($\approx0.2$B params), and (iii) DINOv2~\cite{oquab2023dinov2} ($\approx22$M params).  
Moreover, we compare against other distillation-based approaches: \rev{DRKD \cite{ni2023drkd}}, OFA \cite{hao2023oneforall}, knowledge distillation at the output space via KL divergence (KL) \cite{pmlrv139touvron21a} and its combination with embedding space distillation via MSE (MSE + KL) \cite{fitnets}, open-vocabulary detectors (VILD) \cite{vild2021}, \rev{vision-language grounding models (GLIP) \cite{li2022glip}, similarity-based feature distillation SKDF \cite{ma2023skdf}}, and the most relevant personalization method to date (AuXFT).
\ec[]{Finally, we perform ablations on the student model including YOLOv11n~\cite{yolo11_ultralytics} ($\approx2.6$M params) and RT-DETR-l~\cite{lv2024rtdetrv2} ($\approx45$M params).}

\paragraph{Metrics.}
For consistency and fair comparison, we follow the same implementation details as in AuXFT, except when distilling from the AuXFT-pretrained model. In that case, the number of distillation epochs is reduced from 50 to 20, as the convergence time decreases.
 \revv{We report standard end-to-end mAP$^{50\text{-}95}$ for PerSeg and POD, and retrieval accuracy for CORe50 and iCubWorld following AuXFT~\cite{barbato2024crossarchitectureauxiliaryfeaturespace}, where a prediction is considered correct if the target object is retrieved among the top-ranked detector proposals. Few-shot personalized detection is evaluated under 1- and 5-shot regimes.}

\begin{figure*}[t]
    \centering
    \begin{subfigure}[t]{.33\textwidth}
        \includegraphics[trim=1cm 0cm 0cm 1cm,clip,width=0.9\linewidth]{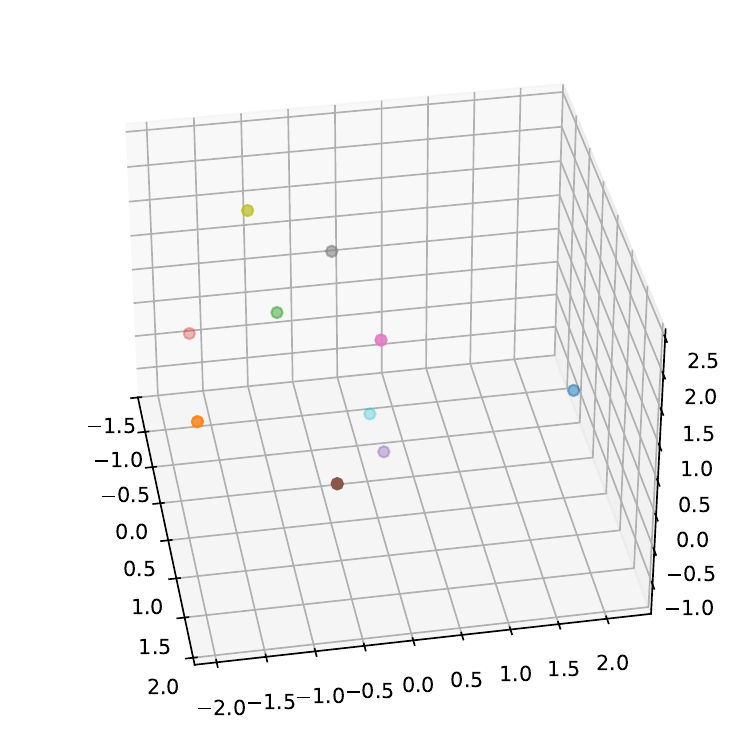}
        \subcaption{3D reference points \\ (proxy for $u'_i$).} \label{fig:emb:refence}
    \end{subfigure}%
    \begin{subfigure}[t]{.33\textwidth}
        \includegraphics[width=0.9\linewidth]{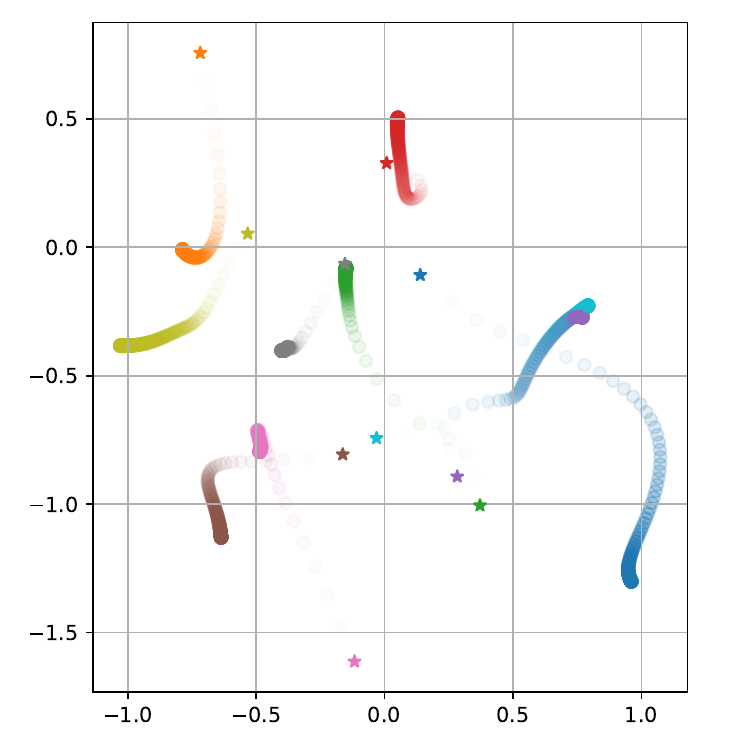}
        \subcaption{2D points being optimized (proxy for $f'_{A,i}$).} 
        \label{fig:emb:evolution}
    \end{subfigure}%
    \begin{subfigure}[t]{.33\textwidth}
        \includegraphics[width=0.9\linewidth]{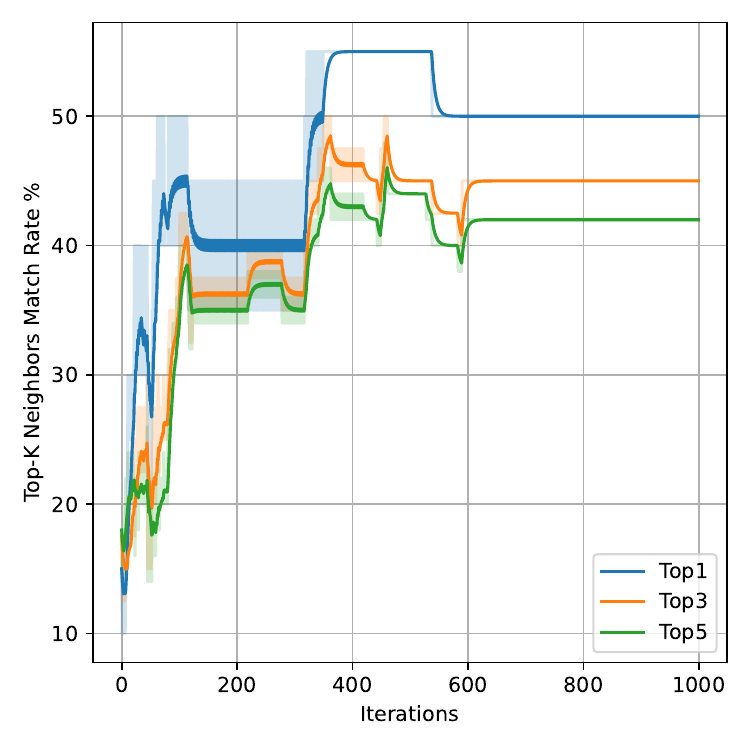}
        \subcaption{k-NN match rate.} 
        \label{fig:emb:topk_match}
    \end{subfigure}
    \caption{\textbf{Effect of $\mathcal{L}_{\mathrm{emb}}$ on a set of ten 2D points optimized with respect to 3D ones.} \textbf{(a)} 3D reference points, proxy for the teacher embeddings $u_i'$. %
    \textbf{(b)} Evolution of the 2D points (proxy for student embeddings $f'_{A,i}$) updated via $\mathcal{L}_{\mathrm{emb}}$ from the 3D reference points $u_i'$. $\star$ marks the original location, timesteps increase with color saturation. \textbf{(c)} Percent rate of 2D top-$k$ nearest neighbors (k-NN) that match those of the reference 3D distribution.}
    \label{fig:embedding}
\end{figure*}

\subsection{Analyses}\label{sec:analyses}

\paragraph{Embedding Loss.}
Fig.~\ref{fig:embedding} illustrates the impact of our relational embedding loss $\mathcal{L}_{\mathrm{emb}}$ on preserving the geometric structure of the teacher features onto student ones. We consider a toy setup, where we optimize the coordinates of a set of 2D points {(proxies for student embeddings $f'_{A,i}$)} to match the pairwise distance distribution of a fixed 3D reference configuration (proxy for teacher embeddings $u'_i$).
Fig.~\ref{fig:emb:refence} shows the 3D reference points sampled uniformly in space. 
Fig.~\ref{fig:emb:evolution} visualizes the optimization trajectory in the 2D space: each path starts from an initial position (marked by $\star$) and progressively adapts under the guidance of $\mathcal{L}_{\mathrm{emb}}$, preserving local neighborhood relationships while globally rearranging to approximate the original distance geometry (color saturation indicates temporal evolution).
Fig.~\ref{fig:emb:topk_match} quantitatively tracks the alignment between neighborhood structures. We report the percentage rate of 2D top-$k$ nearest neighbors that match those of the reference 3D distribution. We observe a steady improvement across all $k$ with a rapid convergence.
This analysis confirms the role of relational supervision in regularizing the student's embedding space beyond pointwise alignment, ultimately promoting feature diversity and semantic separability.

\begin{figure}[t]
    \centering
    \includegraphics[trim=0.2cm 0cm 0cm 0cm,clip,width=0.67\linewidth]{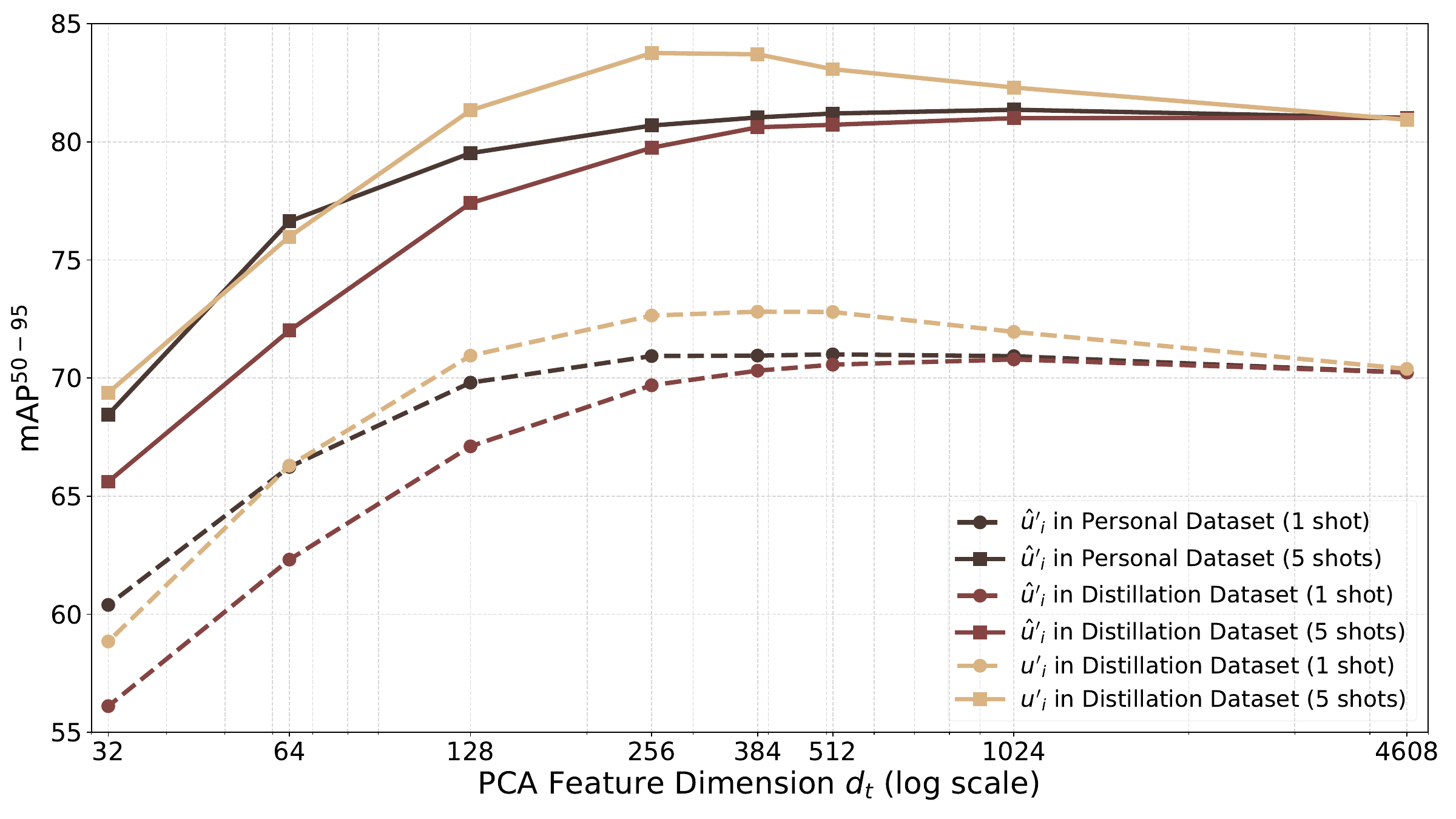}
    \caption{\fb[]{\textbf{mAP$^{50-95}$ at different PCA dimensionality.} %
    Average score} across all evaluation datasets varying feature dimension $d_t$.}
    \label{fig:pca-plot}
\end{figure}

\paragraph{Impact of Teacher Feature Dimensionality.}
Fig.~\ref{fig:pca-plot} evaluates the role of teacher supervision dimensionality ($d_t$) by comparing features: (i) of the personal datasets $\mathcal{D}_f$ (as an upper bound), (ii) of the distillation dataset $\mathcal{D}_c$ unnormalized and (iii) PCA normalized (additional details and validation in \ec[]{Appendix}).
Overall, performance increases steadily up to $d_t=512$, which strikes a good balance between compactness and expressivity. After that, it plateaus or slightly degrades up to the maximum dimension $d = 4608$.  
This indicates that most of the relevant semantics are preserved in a compact subspace, and that further dimensions may introduce redundancy or noise.
\revv{We further analyze modality contributions to the retained PCA components in the Appendix.}
Normalized features $u_i^\prime$ consistently outperform their unnormalized counterparts $\hat{u}_i^\prime$ (especially in the 1-shot setting), obtaining better performance even if compared to personal datasets' features.

\begin{figure}[t]
    \centering
    \resizebox{\textwidth}{!}{%
        \begin{subfigure}{.33\linewidth}
            \includegraphics[width=\linewidth]{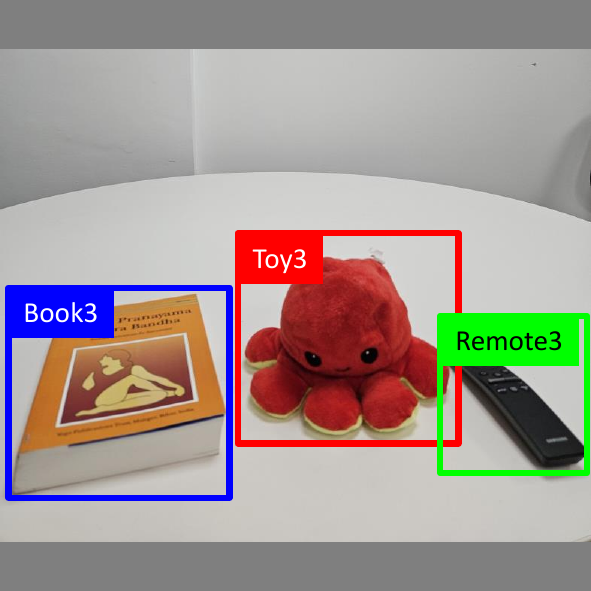}
        \end{subfigure}~%
        \begin{subfigure}{.33\linewidth}
            \includegraphics[width=\linewidth]{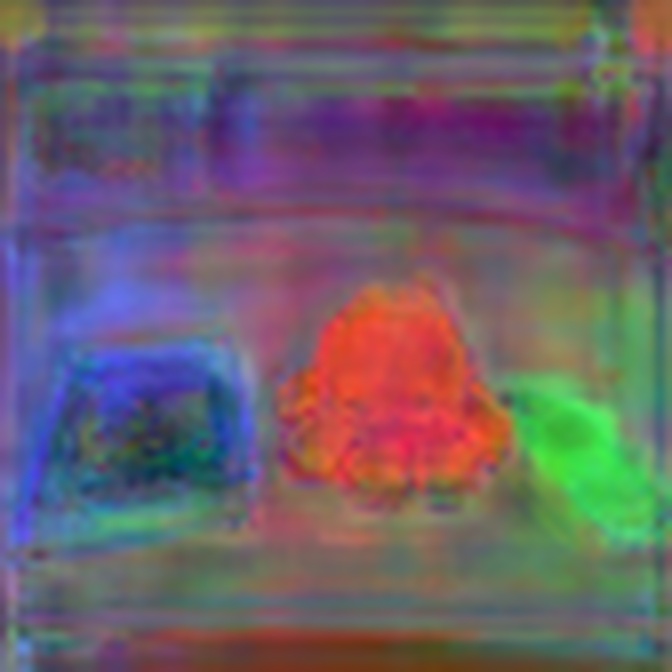}
        \end{subfigure}~~~~~~~%
         \begin{subfigure}{.33\linewidth}
            \includegraphics[width=\linewidth]{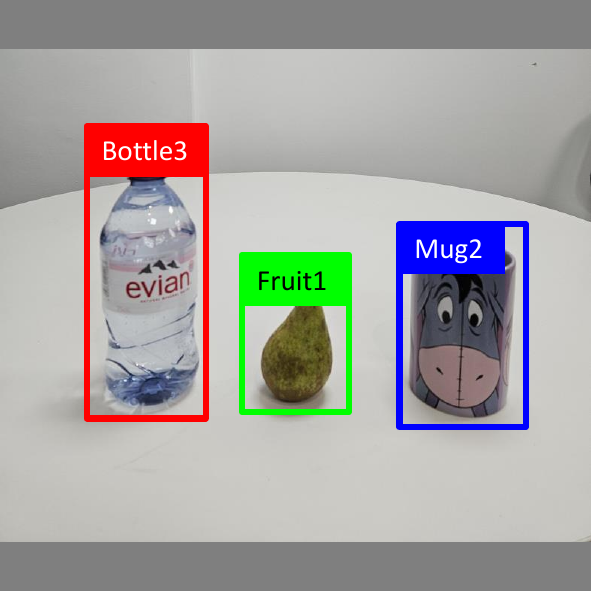}
        \end{subfigure}~%
        \begin{subfigure}{.33\linewidth}
            \includegraphics[width=\linewidth]{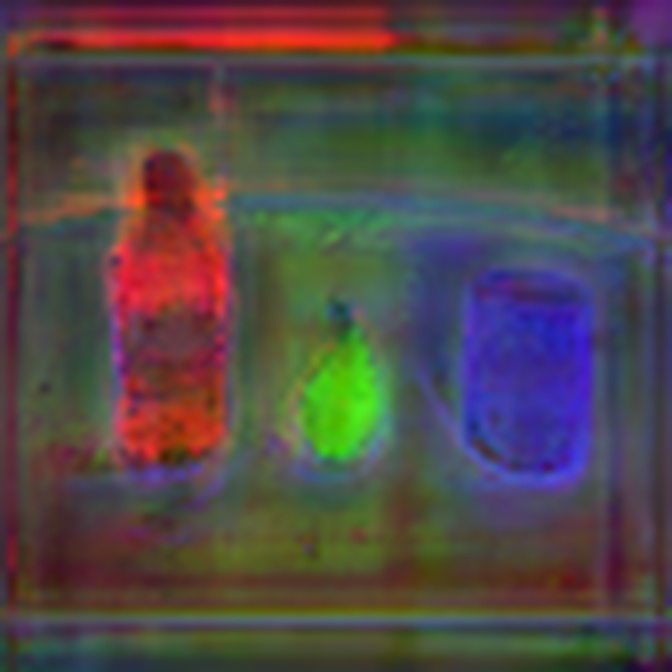}
        \end{subfigure}~~~~~~~%
         \begin{subfigure}{.33\linewidth}
            \includegraphics[width=\linewidth]{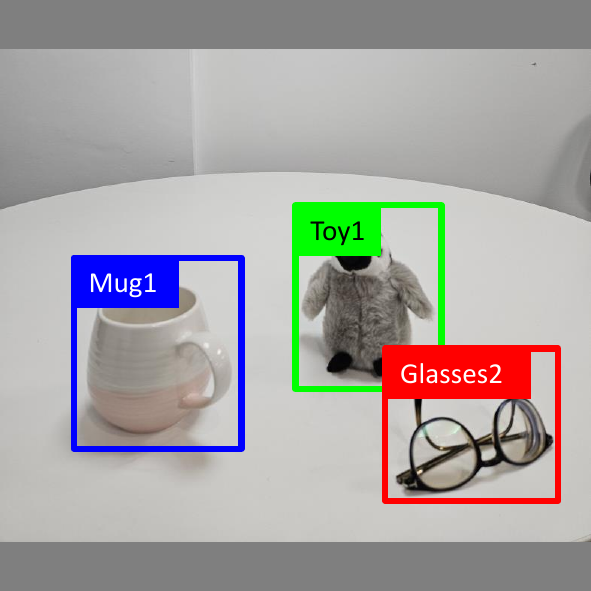}
        \end{subfigure}~%
        \begin{subfigure}{.33\linewidth}
            \includegraphics[width=\linewidth]{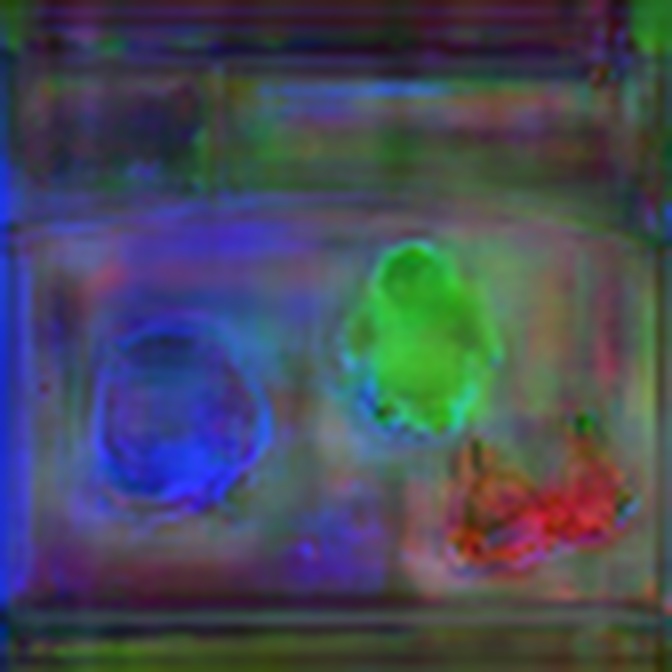}
        \end{subfigure}%
        }%
    \caption{
    \textbf{Feature similarity between the $F_A$ embeddings and the teacher target $u'_i$} encoded in the R/G/B channels, one for each object in input scenes from POD.}
    \label{fig:feat_sim}
\end{figure}

\paragraph{Feature Similarity.}
Fig.~\ref{fig:feat_sim} shows %
the similarity between the teacher vision-language descriptors $u'_i$ of the three objects in each scene and the features output by the student, $F_{A}$. %
In each bottom image, the three similarity matrices are visualized in R/G/B channels, showing that MOCHA successfully maps sparse VLM object-level features (inside object boxes) into the dense features of an efficient detector.

\begin{table*}[t]
    \centering \small
    \setlength{\tabcolsep}{2.5pt}
        \caption{\textbf{Oracle results using \ec[]{ground truth bounding box coordinates} on personal datasets.} Top block: performance when using frozen teacher embeddings directly in the personalization stage. Bottom block: performance of \ec[]{student} distilled with the teacher signals.}
    \label{tab:oracle-results}
    \resizebox{\textwidth}{!}{%
    \begin{tabular}{p{4.7cm}p{1.5cm} ccccccc c}
        \toprule
          \multirow{2}{*}{\textbf{Teacher}} & \multirow{2}{*}{\textbf{Student}} &
          \multicolumn{1}{c}{\textbf{PerSeg}} & \multicolumn{2}{c}{\textbf{POD}} & \multicolumn{2}{c}{\textbf{CORe50}} & \multicolumn{2}{c}{\textbf{iCubWorld}} & \multirow{2}{*}{\textbf{Avg}} \\
        & & {\sc 1 shot} & {\sc 1 shot} & {\sc 5 shot} & {\sc 1 shot} & {\sc 5 shot} & {\sc 1 shot} & {\sc 5 shot} & \\
        \cmidrule(lr){1-1}\cmidrule(lr){2-2}\cmidrule(lr){3-3}\cmidrule(lr){4-5}\cmidrule(lr){6-7}\cmidrule(lr){8-9}\cmidrule(lr){10-10}
 DINO & -- & 90.5 \pms{2.7} & 55.1 \pms{4.0} & 66.8 \pms{0.0} & 34.8 \pms{7.2} & 47.9 \pms{9.8} & 50.6 \pms{2.9} & 73.5 \pms{2.2} & 59.9 \\
CLIP ($z_{V,i}$) & -- & 95.7 \pms{1.4} & 64.5 \pms{3.5} & 77.8 \pms{0.0} & 58.1 \pms{4.6} & 79.3 \pms{4.4} & 55.5 \pms{3.2} & 77.9 \pms{2.5} & 72.7 \\
LLaVa ($h_i$) & -- & 82.2 \pms{3.3} & 59.0 \pms{4.1} & 72.7 \pms{0.5} & 44.1 \pms{3.4} & 63.1 \pms{4.2} & 46.7 \pms{3.1} & 67.6 \pms{2.7} & 62.2 \\
LLaVa ($h_i$) + CLIP ($z_{V,i}, \gamma = 1$) & -- & 93.9 \pms{1.8} & 66.4 \pms{3.8} & 80.7 \pms{0.2} & 59.4 \pms{4.6} & 80.4 \pms{4.3} & 58.0 \pms{3.2} & 80.6 \pms{2.4} & 74.2 \\
LLaVa ($h_i$) + CLIP ($z_{V,i}$) & -- & 95.1 \pms{1.7} & 70.0 \pms{3.4} & 80.6 \pms{0.0} & 57.0 \pms{4.0} & 79.7 \pms{3.7} & 56.4 \pms{3.6} & 81.2 \pms{2.5} & \textbf{74.3} \\
\midrule
-- & YOLO\fb[]{v8n} & 69.6 \pms{3.4} & 35.4 \pms{3.4} & 41.3 \pms{0.0} & 31.5 \pms{5.5} & 42.8 \pms{7.8} & 33.3 \pms{2.7} & 48.6 \pms{2.4} & 43.2 \\
CLIP ($z_{V,i}$) & YOLO\fb[]{v8n} & 75.5 \pms{3.5} & 40.9 \pms{3.2} & 51.7 \pms{0.0} & 35.0 \pms{3.9} & 47.3 \pms{5.7} & 40.6 \pms{2.9} & 60.0 \pms{2.3} & {50.7}  \\
LLaVa ($h_i$) & YOLO\fb[]{v8n} & 73.0 \pms{3.6} & 37.1 \pms{4.2} & 47.6 \pms{0.0} & 33.9 \pms{3.7} & 45.9 \pms{5.4} & 37.4 \pms{3.1} & 55.8 \pms{2.6} & {48.0} \\
LLaVa ($h_i$) + CLIP ($z_{V,i}$) & YOLO\fb[]{v8n} & 82.0 \pms{3.4} & 45.7 \pms{3.5} & 58.1 \pms{0.0} & 34.5 \pms{4.1} & 47.8 \pms{5.9} & 42.9 \pms{3.1} & 62.7 \pms{2.2} & \textbf{54.0} \\
        \bottomrule 
    \end{tabular}%
    }%
\end{table*}

\subsection{Main Results}
We begin our evaluation in the oracle setup, where models are assessed using {ground truth bounding box coordinates} (one per evaluation) during the few-shot personalization stage. 
This setting removes the detection head and focuses solely on the classification capability of each approach, allowing us to compare different teacher models and types of supervision embeddings (linguistic, visual, and multimodal).
Tab.~\ref{tab:oracle-results} is organized into two blocks. The top block reports results obtained by using the teacher embeddings directly during personalization, while the bottom block shows the performance of YOLO students distilled with the same supervision signals.
In the first block, visual teachers show strong performance. Notably, {CLIP} outperforms {DINO} (used in AuXFT). On the other hand, {LLaVa}'s features ($h_i$) perform worse, but combining them with CLIP features ($z_{V,i}$) --- \ie, using $u_i$ for supervision --- yields notable gains at no added cost (as CLIP is already part of LLaVa), demonstrating the complementarity of visual and textual cues.
Normalizing the visual features also leads to a slight improvement over their unnormalized version ($\gamma = 1$).
In the second block, we evaluate distillation using either visual supervision (CLIP, $z_{V,i}$), textual supervision (LLaVa, $h_i$), or their combination (LLaVa, $u_i$). All models are distilled through MOCHA supervision, showing consistent gains of our approach over the non-distilled baseline (YOLO only). Still, using a single modality alone proves less effective than leveraging full multimodal supervision.
\ec[]{Note that in Tab.~\ref{tab:oracle-results} we evaluate using the ground truth boxes, so the retrieval accuracy corresponds to standard detection accuracy, with each image containing a single annotated instance. In Tab.~\ref{tab:comparisons}, instead, the retrieval protocol \cite{barbato2024crossarchitectureauxiliaryfeaturespace} allows multiple candidate boxes per image; we denote an object as correct if \textit{any} predicted box matches the target. As this metric accepts multiple candidates, the scores in Tab.~\ref{tab:comparisons} are naturally slightly higher than those in Tab.~\ref{tab:oracle-results}.}
\rev{These results indicate that improvements do not stem merely from using a stronger multimodal teacher, but from enforcing structured relational alignment in the embedding space.}

\begin{table*}[t]
    \centering \small
    \setlength{\tabcolsep}{2.5pt}
        \caption{\textbf{Few-shot personalized detection results on personal datasets.} First row: \ec[]{student} baseline with no distillation. First block: MOCHA compared against prior state-of-the-art supervision methods. Second block: evaluation of different MOCHA variants.}
    \label{tab:comparisons}
    \resizebox{\textwidth}{!}{%
    \begin{tabular}{p{1.7cm}p{3.1cm}p{1.5cm} ccccccc c}
        \toprule
         \multirow{2}{*}{\textbf{Teacher}} & \multirow{2}{*}{\textbf{Supervision}} & \multirow{2}{*}{\textbf{Student}}
         & \multicolumn{1}{c}{\textbf{PerSeg}} 
         & \multicolumn{2}{c}{\textbf{POD}} 
         & \multicolumn{2}{c}{\textbf{CORe50}} 
         & \multicolumn{2}{c}{\textbf{iCubWorld}} 
         &\multirow{2}{*}{\textbf{Avg}}  \\
        & & & {\sc 1 shot} & {\sc 1 shot} & {\sc 5 shot} & {\sc 1 shot} & {\sc 5 shot} & {\sc 1 shot} & {\sc 5 shot} & \\
        \cmidrule(lr){1-1}\cmidrule(lr){2-2}\cmidrule(lr){3-3}\cmidrule(lr){4-4}\cmidrule(lr){5-6}\cmidrule(lr){7-8}\cmidrule(lr){9-10}\cmidrule(lr){11-11}
-- & -- & YOLO\fb[]{v8n} & 41.2 \pms{2.6} & 23.6 \pms{2.4} & 30.4 \pms{0.0} & 57.8 \pms{6.3} & 67.3 \pms{7.7} & 51.2 \pms{3.0} & 68.4 \pms{2.4} & 48.6 \\
\midrule
DINO & AuXFT & YOLO\fb[]{v8n} & 48.8 \pms{3.4} & 31.5 \pms{2.7} & 38.8 \pms{0.0} & \underline{58.8} \pms{6.4} & \underline{69.3} \pms{6.0} & 55.0 \pms{3.0} & 74.5 \pms{2.5} & 53.8\\
CLIP ($z_{V,i}$) & ViLD & YOLO\fb[]{v8n} & 44.4 \pms{3.2} & 24.8 \pms{2.8} & 33.3 \pms{0.0} & 54.6 \pms{5.2} & 64.9 \pms{4.9} & 57.7 \pms{3.0} & 70.4 \pms{2.4} & 50.9\\
LLaVa ($u_i$) & KL div. & YOLO\fb[]{v8n} & 43.7 \pms{2.9} & 23.4 \pms{2.5} & 27.8 \pms{0.0} & 54.4 \pms{5.0} & 65.2 \pms{5.0} & 54.7 \pms{3.0} & 69.5 \pms{2.3} & 49.0 \\
LLaVa ($u_i$) & KL div. + MSE & YOLO\fb[]{v8n} & 50.1 \pms{3.0} & 27.6 \pms{2.4} & 30.5 \pms{0.0} & 53.3 \pms{4.5} & 65.7 \pms{4.6} & 61.1 \pms{2.9} & \textbf{78.5} \pms{1.9} & 52.1\\
LLaVa ($u_i$) & OFA & YOLO\fb[]{v8n} & 47.2 \pms{3.4} & 23.3 \pms{2.6} & 28.0 \pms{0.0} & 55.3 \pms{4.3} & 63.6 \pms{4.3} & 60.4 \pms{2.5} & 72.5 \pms{2.4} & 50.1\\
LLaVa ($u_i$) & GLIP & YOLO\fb[]{v8n} & 52.1 \pms{2.8} & 27.3 \pms{3.0} & 32.7 \pms{0.0} & 58.7 \pms{5.5} & 64.6 \pms{5.7} & 61.2 \pms{2.4} & 71.1 \pms{2.3} & 52.2 \\
LLaVa ($u_i$) & SKDF & YOLO\fb[]{v8n} & 47.0 \pms{2.9} & 25.8 \pms{2.6} & 28.4 \pms{0.0} & 58.4 \pms{4.8} & 67.3 \pms{5.6} %
& \underline{62.7} \pms{2.9} & 74.1 \pms{2.3} & 52.0 \\ 
LLaVa ($u_i$) & DRKD & YOLO\fb[]{v8n} & 53.5 \pms{2.7} & 24.3 \pms{3.0} & 24.7 \pms{0.0} & 57.7 \pms{5.8} & 64.8 \pms{6.2} & 59.7 \pms{2.7} & 69.7 \pms{2.4} & 50.6 \\
    \midrule
CLIP ($z_{V,i}$) & MOCHA (visual) & YOLO\fb[]{v8n} & 53.0 \pms{2.8} & 26.7 \pms{3.1} & 37.5 \pms{0.0} & 57.2 \pms{4.7} & 67.2 \pms{4.8} & 62.6 \pms{2.8} & 76.1 \pms{2.3} & {54.6}\\
LLaVa ($h_i$) & MOCHA (text) & YOLO\fb[]{v8n} & 
52.3 \pms{2.8} & 28.8 \pms{2.6} & 33.7 \pms{0.0} & 56.2 \pms{4.7} & 67.1 \pms{4.7} & \textbf{62.8} \pms{2.8} & 76.8 \pms{2.0} & {54.0}\\
LLaVa ($u_i$) & \textbf{MOCHA (COCO)} & YOLO\fb[]{v8n} & \underline{55.4} \pms{3.3} & \underline{33.9} \pms{2.8} & \underline{38.9} \pms{0.0} & 55.8 \pms{4.7} & 65.4 \pms{4.7} & 60.6 \pms{3.3} & \underline{77.6} \pms{2.5} & \underline{56.0}\\
LLaVa ($u_i$) & \textbf{MOCHA (AuXFT)} & YOLO\fb[]{v8n} & \textbf{59.1} \pms{3.3} & \textbf{36.3} \pms{3.3} & \textbf{45.9} \pms{0.0} & \textbf{60.9} \pms{4.4} & \textbf{70.6} \pms{4.5} & {61.4} \pms{3.1} & 77.0 \pms{2.3} & \textbf{58.7} \\
        \bottomrule
    \end{tabular}%
   }%
\end{table*}

\paragraph{Comparisons.}
We evaluate the complete approach, including both backbone and detection head, in the standard few-shot personalization setting (\ec[]{predicting bounding box coordinates}).
Tab.~\ref{tab:comparisons} compares MOCHA with recent approaches for personalized detection.
\rev{
To ensure fair comparisons, we re-implemented representative distillation objectives within our pipeline, modifying only the supervision loss while keeping teacher, student, crops, PCA compression, and personalization protocol fixed.
We consider output-space transfer (KL, KL+MSE), cross-architecture feature projection (OFA), similarity/feature regression objectives (GLIP, SKDF), and relational matching (DRKD) as alternative supervision paradigms in our multimodal region-level setting.
}
\rev{
Overall, KL-based and OFA-style supervision provide limited gains, while similarity- and relation-based objectives (GLIP, SKDF, DRKD) improve performance but remain below MOCHA.
This suggests that neither pointwise regression nor relation matching alone is sufficient: combining local alignment with explicit relational regularization is key to effectively transferring multimodal semantics.
}
ViLD struggles in highly personalized scenarios, despite its open-vocabulary capabilities. AuXFT achieves the strongest average performance among prior methods, but still underperforms in more challenging domains (\eg, PerSeg and POD), where fine-grained semantic alignment is critical.
MOCHA surpasses all competing approaches, showing consistently strong results across all datasets and settings, with a +10.1 average improvement over the \ec[]{YOLOv8n} baseline and +4.9 over the strongest competitor, AuXFT.
When initialized from COCO-pretrained weights, MOCHA already exceeds prior state-of-the-art methods; starting from AuXFT-pretrained weights yields further gains, confirming the compatibility and effectiveness of the initialization.
MOCHA introduces only minimal computational overhead: on YOLOv8n on PerSeg, it adds just $3$ ms/image, corresponding to a $\sim$10\% relative increase (see Appendix for details).

\begin{figure}[t]
    \centering
    \resizebox{\textwidth}{!}{%
    \begin{subfigure}{\linewidth}
        \rotatebox{90}{~~~~~YOLOv8n}
        \begin{subfigure}{.3\textwidth}
            \includegraphics[width=\textwidth]{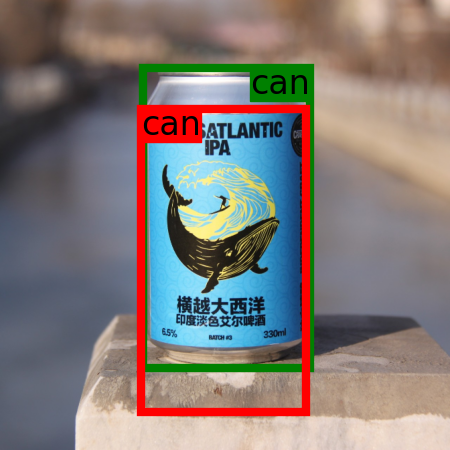}
        \end{subfigure}
        \begin{subfigure}{.3\textwidth}
            \includegraphics[width=\textwidth]{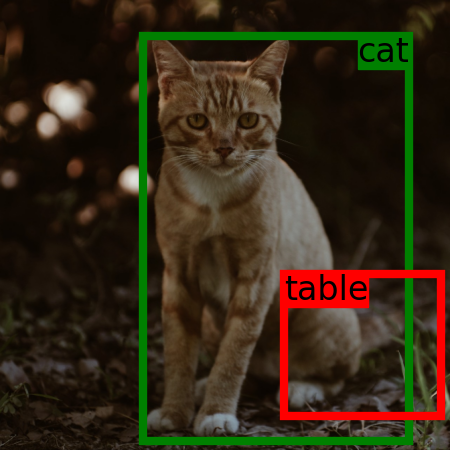}
        \end{subfigure}
        \begin{subfigure}{.3\textwidth}
            \includegraphics[width=\textwidth]{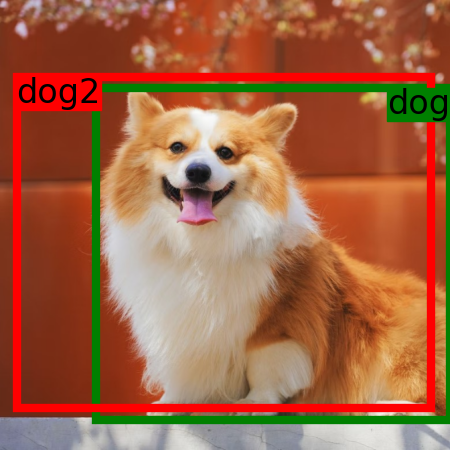}
        \end{subfigure}
    \end{subfigure}
    \begin{subfigure}{\linewidth}
        \rotatebox{90}{~~~~~~~~AuXFT}
        \begin{subfigure}{.3\textwidth}
            \includegraphics[width=\textwidth]{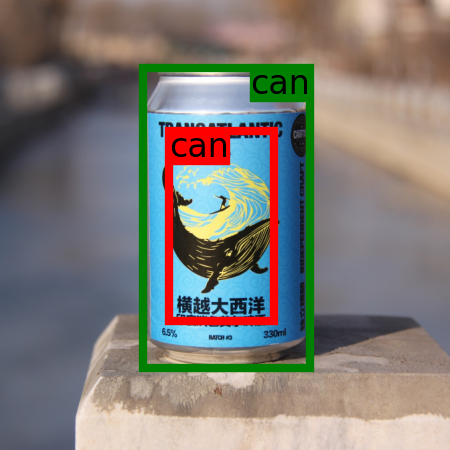}
        \end{subfigure}
        \begin{subfigure}{.3\textwidth}
            \includegraphics[width=\textwidth]{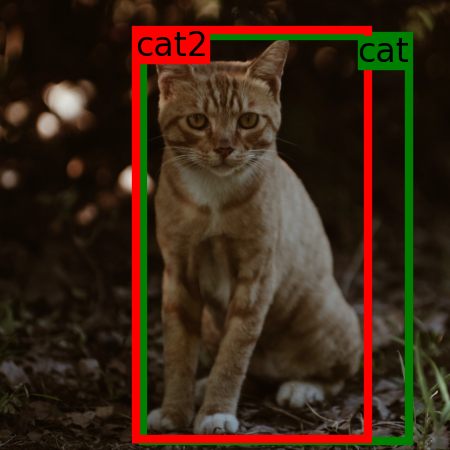}
        \end{subfigure}
        \begin{subfigure}{.3\textwidth}
            \includegraphics[width=\textwidth]{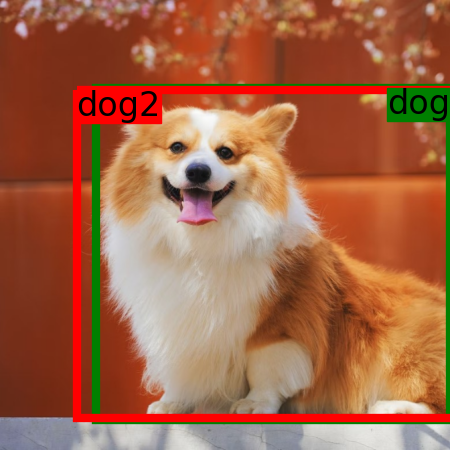}
        \end{subfigure}
    \end{subfigure}%
    }%
    \\
    \resizebox{\textwidth}{!}{%
    \begin{subfigure}{\linewidth}
        \rotatebox{90}{~~~~~~~~~~~OFA}
        \begin{subfigure}{.3\textwidth}
            \includegraphics[width=\textwidth]{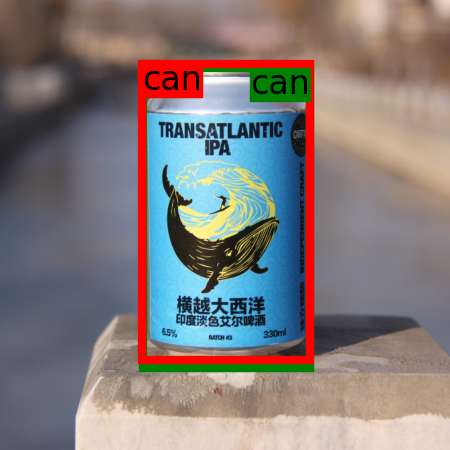}
        \end{subfigure}
        \begin{subfigure}{.3\textwidth}
            \includegraphics[width=\textwidth]{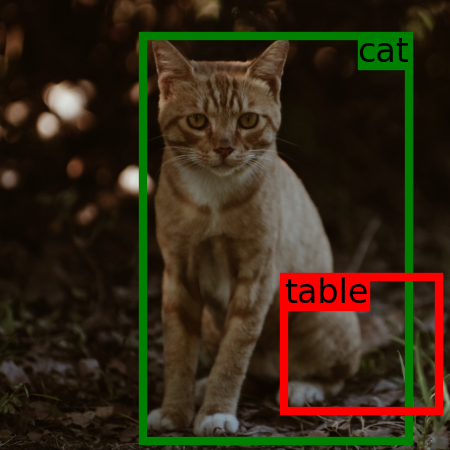}
        \end{subfigure}
        \begin{subfigure}{.3\textwidth}
            \includegraphics[width=\textwidth]{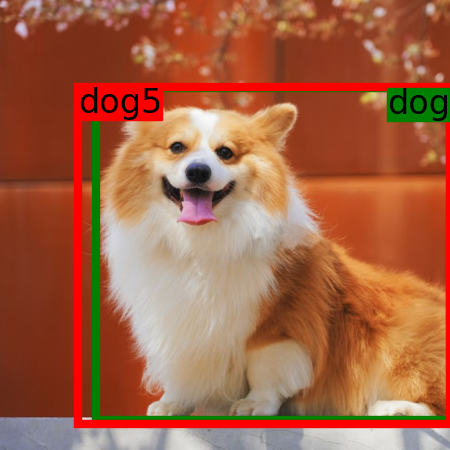}
        \end{subfigure}
    \end{subfigure}
    \begin{subfigure}{\linewidth}
        \rotatebox{90}{~~~~~~~MOCHA*} 
        \begin{subfigure}{.3\textwidth}
            \includegraphics[width=\textwidth]{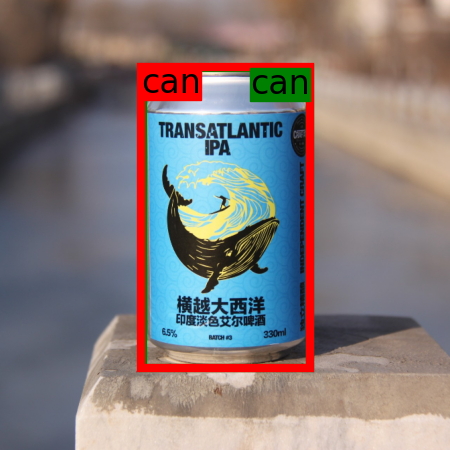}
        \end{subfigure}
        \begin{subfigure}{.3\textwidth}
            \includegraphics[width=\textwidth]{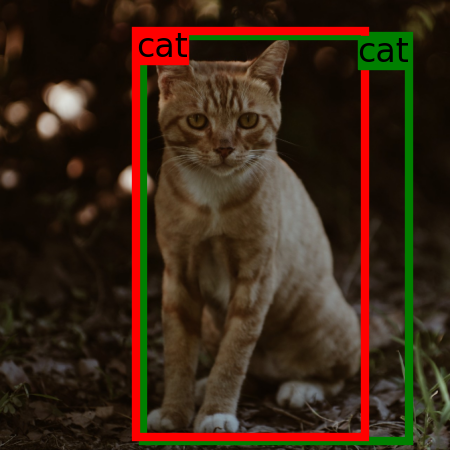}
        \end{subfigure}
        \begin{subfigure}{.3\textwidth}
            \includegraphics[width=\textwidth]{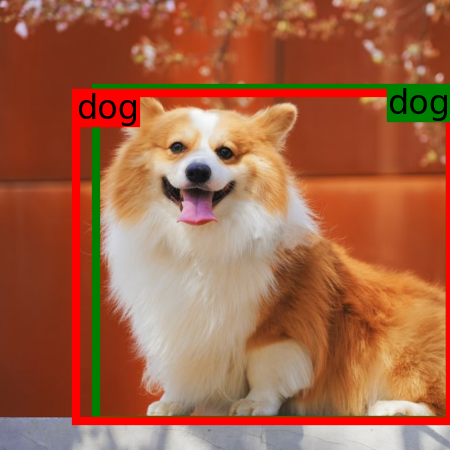}
        \end{subfigure}
    \end{subfigure}%
    }%
    \caption{\textbf{Qualitative results on PerSeg.} Ground truth in green, prediction with the highest confidence in red (class names shown refer to personal class labels in PerSeg). YOLOv8n refers to baseline. *: refers to MOCHA (AuXFT). %
    }
    \label{fig:quali}
\end{figure}

\paragraph{Qualitative Results.}
Fig.~\ref{fig:quali} presents a qualitative comparison 
on the PerSeg dataset, featuring three sample images: a \textit{can}, a \textit{cat}, and a \textit{dog}. Green bounding boxes indicate ground truth annotations, while red boxes show the highest-confidence predictions. 
{MOCHA (AuXFT)} consistently demonstrates superior performance across all samples, with correct class predictions and precise bounding boxes. In contrast, competing methods exhibit occasional misclassifications and less accurate detections.  
\ec[]{These qualitative examples reflect the trends observed in the quantitative results, with MOCHA offering more reliable predictions.}

\subsection{Ablation Studies}
\paragraph{Generalization to Other Student Models.}
Tab.~\ref{tab:abl-student} reports an ablation on the student model. Results show that MOCHA consistently improves few-shot detection across all student architectures, demonstrating strong generalization beyond the YOLOv8n model used in the main experiments.
The largest gains are obtained with YOLOv8n, where the combination of MOCHA's region-level supervision and YOLO's multi-scale feature hierarchy yields substantial improvements over both the baseline and AuXFT. 
Upgrading the student to YOLOv11n preserves this trend, albeit with smaller margins. This suggests that newer backbones, while stronger on their own, leave less room for improvement but still benefit from MOCHA supervision.
On the other hand, the transformer-based RT-DETR-l shows the smallest but still consistent gains. 
\ec[]{This is because RT-DETR-l is much larger ($\approx 10$ times) and architecturally very different from YOLO detectors, making the translation module less effective and limiting its suitability for on-device use.
Despite this, MOCHA still provides consistent gains across datasets, confirming its robustness and ability to offer complementary supervision across detector families.}
\rev{
Nevertheless, performance variations across architectures reflect their different inductive biases and the magnitude of the semantic gap with respect to the VLM teacher. 
CNN-based detectors (\eg, YOLO variants) benefit more from multimodal alignment, while transformer-based detectors (\eg, RT-DETR), being architecturally closer to the teacher, exhibit smaller but consistent improvements.
}

\paragraph{Components Design.} Tab.~\ref{tab:abl-results} reports an extensive ablation study, isolating the contributions of key design and training components of MOCHA across the PerSeg and iCubWorld datasets.
We first assess the impact of the PCA dimensionality $d_t$ used to compress teacher embeddings. As previously highlighted in Fig.~\ref{fig:pca-plot}, compressing the teacher's embedding improves performance compared to using full-resolution features.
A PCA dimension of $d_t = 512$ \ec[]{represents the best} balance between compactness and expressiveness. \ec[]{For this reason, we selected it as the embedding size}.
More aggressive compression (\eg, $256$ or $384$) leads to reduced performance, indicating a loss of relevant semantic content\ec[]{, while larger ones (\eg, $1024$) increase computational cost without significant gains.}
Next, we examine the role of core MOCHA components. Removing the embedding alignment loss ($\mathcal{L}_{\mathrm{emb}}$) degrades accuracy on both datasets, underscoring its importance for preserving global teacher structure. Disabling attention mechanisms in the translator ($\mathrm{t}_S$ {no attn.}) also leads to a consistent drop, suggesting that attention layers play a key role in adapting multimodal features \revv{A further linear-translator ablation is found in the Appendix.} %
Likewise, removing the translator's final layer ($\mathrm{t}_S$ {no ffn}) reduces performance. Enabling data augmentation (standard YOLO augmentations only varying color, $\mathcal{D}_c$ {augm.}) lowers robustness, while manipulating learning rates — doubling them for encoder or decoder separately — slightly harms performance. 
Finally, we compare different pretraining strategies. When pretraining is performed on OpenImages only, performance remains below that of MOCHA configurations. 
COCO-pretrained weights yield a substantial boost, but AuXFT-pretrained weights lead to the best results overall, thanks to their strong initial performance and architectural compatibility with MOCHA.

\begin{table*}[t]
\centering
\begin{minipage}[t]{0.49\columnwidth}
    \centering \small
    \setlength{\tabcolsep}{2.5pt} %
        \caption{\ec[]{\textbf{Ablation studies on the student model.} Comparison between AuXFT and MOCHA (AuXFT) supervisions using different student models: YOLOv8n, YOLOv11n, RT-DETR-l.\vspace{1em}}
    } 
    \label{tab:abl-student}
    \resizebox{0.98\columnwidth}{!}{%
    \begin{tabular}{ll ccc}
        \toprule
         \multirow{2}{*}{\textbf{Supervision}} & \multirow{2}{*}{\textbf{Student}} & \multicolumn{1}{c}{\textbf{PerSeg}} & \multicolumn{2}{c}{\textbf{iCubWorld}} \\
        & & {\sc 1 shot} & {\sc 1 shot} & {\sc 5 shot}  \\
        \cmidrule(lr){1-2}\cmidrule(lr){3-3}\cmidrule(lr){4-5}
        -- & YOLOv8n & 41.2 \pms{2.6} & 51.2 \pms{3.0} & 68.4 \pms{2.4} \\ %
        AuXFT & YOLOv8n & {48.8} \pms{3.4} & {55.0} \pms{3.0} & {74.5} \pms{2.5} \\
        \textbf{MOCHA} & YOLOv8n & \textbf{59.1} \pms{3.3} & \textbf{61.4} \pms{3.1} & \textbf{77.0} \pms{2.3} \\
        \midrule
        -- & YOLOv11n & {49.3} \pms{3.8} & {52.7} \pms{4.1} & {72.8} \pms{2.7} \\
        AuXFT & YOLOv11n & {52.8} \pms{3.8} & {53.9} \pms{4.1} & {73.3} \pms{2.4} \\
        \textbf{MOCHA} & YOLOv11n & \textbf{56.2} \pms{3.2} & \textbf{57.5} \pms{2.5} & \textbf{74.1} \pms{2.4} \\
        \midrule
        -- & RT-DETR-l & 43.5 \pms{3.5} & 45.0 \pms{3.8} & 63.5 \pms{2.5} \\
        AuXFT & RT-DETR-l & 45.2 \pms{3.4} & 46.1 \pms{3.7} & 64.0 \pms{2.4} \\
        \textbf{MOCHA} & RT-DETR-l &  \textbf{47.0} \pms{3.1} & \textbf{47.8} \pms{3.2} & \textbf{64.4} \pms{2.3}  \\
         \bottomrule
    \end{tabular}%
    }%
\end{minipage}
\hfill
\begin{minipage}[t]{0.49\columnwidth}
    \centering \small
    \setlength{\tabcolsep}{2.5pt} %
        \caption{\textbf{Ablation studies on MOCHA components}.
        Blocks: PCA dimension $d_t$, MOCHA components, pretrainings. %
        \vspace{-0.5em}
        }
    \label{tab:abl-results}
    \resizebox{\columnwidth}{!}{%
    \begin{tabular}{lcc ccc}
        \toprule
         \multirow{2}{*}{\textbf{Pretr.}} & \multirow{2}{*}{{${d_t}$}} & \multirow{2}{*}{\textbf{Case}} & \multicolumn{1}{c}{\textbf{PerSeg}} & \multicolumn{2}{c}{\textbf{iCubWorld}} \\
        & & & {\sc 1 shot} & {\sc 1 shot} & {\sc 5 shot}  \\
        \cmidrule(lr){1-3}\cmidrule(lr){4-4}\cmidrule(lr){5-6}
        OI & -- & (YOLOv8n) & 41.2 \pms{2.6} & 51.2 \pms{3.0} &  68.4 \pms{2.4} \\
        \midrule
        AuXFT & 256 & -- & \underline{56.4} \pms{3.3} & {61.2} \pms{2.6} & 75.7 \pms{2.1} \\
        AuXFT & 384 & -- & 56.2 \pms{3.5} & 59.6 \pms{3.0} & 74.7 \pms{2.0} \\
        AuXFT & 1024 & -- & 53.8 \pms{3.5} & 58.4 \pms{3.1} & \underline{77.0} \pms{2.3} \\
        \midrule
        AuXFT & 512 & \cancel{$\mathcal{L}_{emb}$} & 56.6 \pms{3.3} & 59.8 \pms{2.9} & 74.9 \pms{2.0} \\ 
        AuXFT & 512 & $\mathrm{t}_S$ no attn. & 56.8 \pms{3.3} & 57.5 \pms{3.1} & 73.2 \pms{2.1} \\
        AuXFT & 512 & $\mathrm{t}_S$ no ffn & 55.7 \pms{3.4} & \textbf{61.4} \pms{3.0} & 76.6 \pms{2.3} \\ 
        AuXFT & 512 & $\mathcal{D}_c$ augm. & 53.3 \pms{3.9} & 58.4 \pms{2.9} & 73.4 \pms{2.1} \\
        AuXFT & 512 & $2\times$lr (enc) & 53.6 \pms{2.9} & 59.0 \pms{3.0} & 75.8 \pms{2.5} \\ 
        AuXFT & 512 & $2\times$lr (dec) & 53.9 \pms{2.9} & 59.7 \pms{3.1} & 75.9 \pms{2.5} \\ 
        \midrule
        OI & 512 & (\textbf{MOCHA}) & 53.8 \pms{3.2} & 56.8 \pms{2.7} & 70.4 \pms{2.3} \\
        COCO & 512 & (\textbf{MOCHA}) & {55.4} \pms{3.3} & 60.6 \pms{3.3} & \textbf{77.6} \pms{2.5} \\
        AuXFT & 512 & (\textbf{MOCHA}) & \textbf{59.1} \pms{3.3} & \textbf{61.4} \pms{3.1} & \underline{77.0} \pms{2.3} \\
         \bottomrule
    \end{tabular}%
    }%
\end{minipage}
\vskip -0.5em
\end{table*}

\section{Conclusion}
In this work, we introduced MOCHA, an effective multimodal objects-aware cross-architecture alignment technique that transfers region-level knowledge from large multimodal foundation models into compact vision detectors, enabling robust few-shot personalized object detection.
Across four personal benchmarks (PerSeg, POD, CORe50, and iCubWorld), MOCHA consistently surpasses prior approaches, achieving an average improvement of +10.1 over the YOLOv8n baseline and outperforming the best competitor, AuXFT, by +4.9. 
Our method generalizes well across different student architectures, including YOLOv11n and RT-DETR-l. 
MOCHA transfers knowledge from large multimodal models into compact visual detector which are well suited for edge deployment in resource-constrained scenarios, such as mobile and robotic platforms.

\appendix

\section{Appendix}

In this appendix, we provide additional experiments and analyses to further validate our methodology.
We begin with a statistical validation of MOCHA using the Wilcoxon signed-rank test to assess the significance of our findings.
\fb[]{We then present a comprehensive performance evaluation, including additional experiments that vary either the student model or the FSL architecture.}
Next, we outline the implementation details of our experiments and provide pseudocode describing the MOCHA pipeline.
\ec[]{Finally, we detail the PCA curve fitting procedure and report per-dataset performance breakdowns. We present qualitative results and discuss the current limitations of MOCHA.}

\section{Wilcoxon Experiment}
To verify the statistical significance of the results presented in the main document, in Tab.~\ref{tab:wilcoxon} we report the Wilcoxon signed rank metric computed on the mAP achieved by MOCHA and its competitors in the PerSeg and POD datasets. The analysis is performed on the same episodic experiments in Tab.~2 (in the main document), where average scores are reported.
The results strongly confirm the statistical superiority of MOCHA, yielding extremely small $p$-values (often below $10^{-17}$) across the majority of comparisons.

\begin{table}[ht]
\caption{$p$-value of \textbf{Wilcoxon signed-rank metric} on PerSeg/POD 1-shot between MOCHA (AuXFT) and competitors.}
\label{tab:wilcoxon}
\centering \scriptsize
\setlength{\tabcolsep}{2.5pt}
\resizebox{0.5\textwidth}{!}{%
\begin{tabular}{l cc}
\toprule
         \multirow{2}{*}{\textbf{Supervision}} 
         & \multicolumn{1}{c}{\textbf{PerSeg}} & \multicolumn{1}{c}{\textbf{POD}} \\
& {\sc 1 shot} & {\sc 1 shot} \\
\midrule
--                        & $1.9\times 10^{-18}$ & $1.9\times 10^{-18}$ \\
\midrule
AuXFT                       & $4.5\times 10^{-18}$ & $1.1\times 10^{-9}$ \\
ViLD                        & $2.1\times 10^{-18}$ & $1.9\times 10^{-18}$ \\
KL div.                     & $1.9\times 10^{-18}$ & $1.9\times 10^{-18}$ \\
MSE + KL div.              & $2.1 \times 10^{-18}$ & $1.9\times 10^{-18}$\\
OFA                         & $1.9\times 10^{-18}$ & $1.9\times 10^{-18}$ \\
\midrule
MOCHA (visual) & $2.0\times 10^{-18}$ & $1.9\times 10^{-18}$ \\
MOCHA (text)
& $1.9\times 10^{-18}$ & $2.0\times 10^{-18}$ \\
\textbf{MOCHA (COCO)}      & $5.5\times 10^{-13}$ & $7.7\times 10^{-9}$ \\
\bottomrule
\end{tabular}%
}
\end{table}

\section{Performance Evaluation}
\rev{Due to its offline design, distillation between teacher and student models is efficient: with batch size $B=32$, training runs at approximately 330 ms/iteration with a peak memory footprint of $\sim$19.6\,GB.
}
\fb[]{Tab.~\ref{tab:performance} summarizes the key computational properties of each supervision strategy, reporting the number of parameters, training time (in hours), inference throughput (ms/im), maximum VRAM usage, and model size (in MB).}
\fb[]{The table shows how MOCHA requires much less training time than its closest competitor, almost matching the time needed to train the detector alone.
This is, in part, due to the efficient implementation of the MOCHA distillation system. Since we employ deterministic augmentation during training, it is possible to cache the mixed LLAVA-CLIP embedding vectors after the first computation. As a consequence, the overall computational complexity is reduced by several orders of magnitude, yielding a dramatic speedup over a full training.}
\ec[]{At the same time, MOCHA maintains a very lightweight inference footprint, introducing only a minimal overhead compared to AuXFT while providing significantly stronger downstream performance.}

\subsection{End-to-End VLM Personalization without Ground-Truth Boxes}
\revv{We additionally evaluate in Tab.~\ref{tab:direct_vlm_no_gt} frozen VLM features without ground-truth boxes by classifying/reranking the same candidate regions used by MOCHA.
This protocol provides a direct teacher-feature reference while preserving the realistic end-to-end proposal setting.
MOCHA does not use VLM inference at test time.}

\begin{table*}[t]
\caption{\ec[]{\textbf{Performance evaluation.} Comparison of computational cost, memory usage, and inference efficiency across supervision strategies. }}
\label{tab:performance}
\centering \small
\setlength{\tabcolsep}{2.5pt}
\resizebox{\textwidth}{!}{%
\begin{tabular}{l ccc ccc}
\toprule
         \multirow{2}{*}{\textbf{Supervision}} & \multirow{2}{*}{\textbf{Params}} 
         & \multicolumn{2}{c}{\textbf{Pretraining}} & \multicolumn{3}{c}{\textbf{FSL Inference}}\\
         & & {\textbf{Dist. Complexity}} & {\textbf{Time [hours]}} & {\textbf{Time [ms/im]}} & \textbf{VRAM [MB]} & {\textbf{Size [MB]}}\\
\cmidrule(lr){1-1}\cmidrule(lr){2-2}\cmidrule(lr){3-4}\cmidrule(lr){5-7}
--       & $3.2$M  & $O(\bar{N})$ & $10$h & {50.5} \pms{5.3} & $221.5$ & $12.2$\\
AuXFT   & $3.9$M  & $O(\bar{N}\times \bar{H} \times \bar{W})$ & $50$h & 32.6 \pms{5.1} & $288.7$ & $18.4$\\
 \textbf{MOCHA (AuXFT)} & $4.5$M  & $O(\bar{N})$ & $11$h  & 35.2 \pms{4.6} & $352.5$ & $22.8$ \\ 
\bottomrule
\end{tabular}%
}
\end{table*}

\begin{table*}[t]
\centering
\scriptsize
\renewcommand{\arraystretch}{0.9}
\setlength{\tabcolsep}{4pt}
\caption{\revv{End-to-end personalization without ground-truth boxes. Candidate regions are classified/reranked under the same protocol.}}
\label{tab:direct_vlm_no_gt}
\begin{tabular}{l ccc}
\toprule
\multirow{2}{*}{\textbf{Model}} &
\textbf{PerSeg} & \multicolumn{2}{c}{\textbf{POD}} \\
& {\sc 1 shot} & {\sc 1 shot} & {\sc 5 shot} \\
\cmidrule(lr){1-1}\cmidrule(lr){2-2}\cmidrule(lr){3-4}
\multicolumn{4}{l}{\textit{Teacher Upper Bound (Frozen VLM features)}} \\
LLaVa ($h_i$) & 81.9\pms{3.2} & 58.8\pms{4.2} & 72.7\pms{0.5} \\
CLIP ($z_{V,i}$) & 95.1\pms{1.5} & 64.5\pms{3.5} & 77.8\pms{0.1} \\
LLaVa ($h_i$) + CLIP ($z_{V,i}$) & 94.0\pms{1.8} & 66.4\pms{3.8} & 80.7\pms{0.2} \\
LLaVa + CLIP + PCA ($512$) & \textbf{94.3}\pms{1.7} & \textbf{67.8}\pms{3.5} & \textbf{80.1}\pms{0.0} \\
\midrule
\multicolumn{4}{l}{\textit{Student (Efficient Edge Deployment)}} \\
\textbf{MOCHA (YOLOv8n)} & 59.1\pms{3.3} & 36.3\pms{3.3} & 45.9\pms{0.0} \\
\bottomrule
\end{tabular}
\end{table*}

\begin{table*}[t]
\caption{\ec[]{\textbf{Ablation studies on the student model.} Comparison between AuXFT and MOCHA (AuXFT) supervisions using different student models: YOLOv8n, YOLOv11n, YOLOv11s, YOLOv11l.}}
\label{tab:abl-student}
\centering \small
\setlength{\tabcolsep}{2.5pt}
\resizebox{\textwidth}{!}{%
\begin{tabular}{p{1.8cm}p{1.8cm}p{1.2cm} ccccccc c}
        \toprule
         \multirow{2}{*}{\textbf{Supervision}} & \multirow{2}{*}{\textbf{Student}} & \multirow{2}{*}{\textbf{Params}} & \multicolumn{1}{c}{\textbf{PerSeg}} & \multicolumn{2}{c}{\textbf{POD}} & \multicolumn{2}{c}{\textbf{CoRE50}} & \multicolumn{2}{c}{\textbf{iCubWorld}} & \multirow{2}{*}{\textbf{Avg}} \\
        & & & {\sc 1 shot} & {\sc 1 shot} & {\sc 5 shot} & {\sc 1 shot} & {\sc 5 shot} & {\sc 1 shot} & {\sc 5 shot} & \\
        \cmidrule(lr){1-1}\cmidrule(lr){2-2}\cmidrule(lr){3-3}\cmidrule(lr){4-4}\cmidrule(lr){5-6}\cmidrule(lr){7-8}\cmidrule(lr){9-10}\cmidrule(lr){11-11}
        -- & YOLOv8n & $3.2$M & 41.2 \pms{2.6} & 23.6 \pms{2.4} & 30.4 \pms{0.0} & 57.8 \pms{6.3} & 67.3 \pms{7.7} & 51.2 \pms{3.0} & 68.4 \pms{2.4} & 48.6 \\
        AuXFT & YOLOv8n & $3.9$M & 48.8 \pms{3.4} & 31.5 \pms{2.7} & 38.8 \pms{0.0} & {58.8} \pms{6.4} & {69.3} \pms{6.0} & 55.0 \pms{3.0} & 74.5 \pms{2.5} & 53.8\\
        {\textbf{MOCHA}} & YOLOv8n & $4.5$M & \textbf{59.1} \pms{3.3} & \textbf{36.3} \pms{3.3} & \textbf{45.9} \pms{0.0} & \textbf{60.9} \pms{4.4} & \textbf{70.6} \pms{4.5} & \textbf{61.4} \pms{3.1} & \textbf{77.0} \pms{2.3} & \textbf{58.7} \\
        \midrule
        -- & YOLOv11n & $2.6$M & {49.3} \pms{3.8} & 23.7 \pms{3.0} & 30.5 \pms{0.0} & 32.2 \pms{5.8} & 55.3 \pms{7.2} & {52.7} \pms{4.1} & {72.8} \pms{2.7} & 45.2 \\
        AuXFT & YOLOv11n  & $3.3$M & {52.8} \pms{3.8} & 27.3 \pms{2.8} & 37.9 \pms{0.0} & 33.0 \pms{6.0} & 57.0 \pms{6.5} &  {53.9} \pms{4.1} & {73.3} \pms{2.4} & 47.9 \\
        \textbf{MOCHA} & YOLOv11n & $3.9$M & \textbf{56.2} \pms{3.2} & \textbf{29.0} \pms{2.2} & \textbf{38.2} \pms{0.0} & \textbf{58.6} \pms{5.9} & \textbf{66.8} \pms{2.0} & \textbf{57.5} \pms{2.5} & \textbf{74.1} \pms{2.4} & \textbf{54.3} \\
         \midrule
        -- & YOLOv11s & $9.4$M & 50.6 \pms{3.4} & 27.0 \pms{2.8} & 40.4 \pms{0.0} & 42.6 \pms{6.2} & 63.2 \pms{8.3} & 52.9 \pms{3.6} & 72.4 \pms{2.7} & 49.9 \\
        AuXFT & YOLOv11s & $10.1$M & 54.6 \pms{3.4} & 32.7 \pms{3.5} & 42.2 \pms{0.0} & 45.2 \pms{6.5} & 67.8 \pms{7.0} & 54.5 \pms{3.5} & 75.3 \pms{2.3} & 53.2 \\
        \textbf{MOCHA} & YOLOv11s & $10.7$M & \textbf{57.0} \pms{3.4} & \textbf{37.7} \pms{3.1} & \textbf{46.3} \pms{0.0} & \textbf{56.6} \pms{6.7} & \textbf{68.6} \pms{7.2} & \textbf{58.7} \pms{3.0} & \textbf{76.7} \pms{2.5} & \textbf{57.4}\\
         \midrule
        -- & YOLOv11l & $25.3$M & 47.6 \pms{3.6} & 26.3 \pms{3.0} & 36.9 \pms{0.0} & 38.9 \pms{7.6} & 60.5 \pms{8.6} & 49.9 \pms{3.7} & 65.6 \pms{2.6} & 46.5 \\
        AuXFT & YOLOv11l & $26.0$M & 49.4 \pms{3.7} & 32.4 \pms{3.1} & 46.0 \pms{0.0} & 40.1 \pms{7.9} & 63.5 \pms{8.4} & 53.1 \pms{4.1} & 72.2 \pms{2.4} & 51.0 \\
        \textbf{MOCHA} & YOLOv11l & $26.6$M & \textbf{56.3} \pms{3.9} & \textbf{38.8} \pms{2.7} & \textbf{47.6} \pms{0.0} & \textbf{53.8} \pms{6.3} & \textbf{61.0} \pms{7.9} & \textbf{56.8} \pms{3.4} & \textbf{75.0} \pms{2.3} & \textbf{55.6}\\
         \bottomrule
    \end{tabular}%
    }
    \label{tab:abl-student}
\end{table*}

\newpage
\section{Additional Ablation Studies}
\ec[]{In this section we provide further ablation experiments to support our findings.}

\subsection{Student Model}
\fb[]{In Tab.~\ref{tab:abl-student} we report the performance of MOCHA when applied to different YOLO architectures: YOLOv8n \cite{yolov8_ultralytics} (architecture used in the main experiments), YOLOv11n/s/l \cite{yolo11_ultralytics} (more recent and higher-performing versions).
MOCHA consistently outperforms both the baseline and the closest competitor, across all settings and datasets. Interestingly, the magnitude of the improvement tends to decrease with larger models: while lightweight models such as YOLOv8n and YOLOv11n \revv{benefit from substantial gains in the reported metrics}, becoming more moderate for the larger YOLOv11s and YOLOv11l architectures. }
We hypothesize that this behaviour results from the higher representational capacity of large networks, which can reduce the effectiveness of MOCHA's efficiency-oriented distillation strategy.

\subsection{FSL Architecture}
\ec[]{To further assess the robustness of MOCHA's embeddings, Tab.~\ref{tab:abl-fsl} compares the performance of standard YOLOv8n baseline, AuXFT, and MOCHA (AuXFT) varying the architecture used for Few-Shot Learning classification.
We used the same FSL architectures as in \cite{barbato2024crossarchitectureauxiliaryfeaturespace} plus a learned linear probing classifier.
MOCHA consistently provides the best results, across all settings and datasets, demonstrating that the distilled representations remain highly effective regardless of the downstream FSL classifier. 
This confirms MOCHA's ability to generalize reliably across different learning paradigms, from metric-based approaches such as ProtoNet \cite{snell2017prototypical} and SimpleShot \cite{wang2019simpleshotrevisitingnearestneighborclassification} to parametric classifiers such as Linear Probing.}

\begin{table}[t]
\caption{{\textbf{Ablation studies on the FSL architecture.} Comparison between AuXFT and MOCHA (AuXFT) supervisions using different Few-Shot Learners: ProtoNet (AuXFT), ProtoNet, SimpleShot, Linear Probing (single linear with bias; Adam, $lr=0.1$).}}
\label{tab:abl-fsl}
\centering \small
\setlength{\tabcolsep}{2.5pt} 
\resizebox{0.5\textwidth}{!}{%
\begin{tabular}{p{1.8cm}p{2.8cm} cc}
        \toprule
        \multirow{2}{*}{\textbf{Supervision}} & \multirow{2}{*}{\textbf{FSL Architecture}} & \multicolumn{1}{c}{\textbf{PerSeg}} & \multicolumn{1}{c}{\textbf{POD}} \\
        & & {\sc 1 shot} & {\sc 1 shot}  \\
        \cmidrule(lr){1-2}\cmidrule(lr){3-3}\cmidrule(lr){4-4}
        -- & ProtoNet (AuXFT) & 41.2 \pms{2.6} & 23.6 \pms{2.4} \\ %
        AuXFT & ProtoNet (AuXFT) & {48.8} \pms{3.4} & {31.5} \pms{2.7} \\ %
        \textbf{MOCHA} & ProtoNet (AuXFT) & \textbf{59.1} \pms{3.3} & \textbf{36.3} \pms{3.3}\\ %
        \midrule
        -- & ProtoNet ($\ell_2$) & {43.4} \pms{3.4} & {16.5} \pms{2.3} \\
        AuXFT & ProtoNet ($\ell_2$) & {54.1} \pms{3.4} & {31.0} \pms{2.9} \\ %
        \textbf{MOCHA} & ProtoNet ($\ell_2$) & \textbf{62.7} \pms{3.0} & \textbf{32.7} \pms{2.9} \\ %
        \midrule
        -- & SimpleShot & {42.8} \pms{3.1} & {16.4} \pms{2.4} \\ %
        AuXFT & SimpleShot & {53.7} \pms{3.2} & {30.5} \pms{2.8} \\
        \textbf{MOCHA} & SimpleShot & \textbf{61.6} \pms{3.4} & \textbf{33.3} \pms{2.8} \\
        \midrule
        -- & Linear Probing & {~2.9} \pms{1.2} & {1.5} \pms{1.0} \\
        AuXFT & Linear Probing & {22.8} \pms{2.9} & {9.6} \pms{2.0} \\
        \textbf{MOCHA} & Linear Probing & \textbf{42.8} \pms{3.0} & \textbf{19.5} \pms{2.4} \\
         \bottomrule
    \end{tabular}%
    }
\end{table}

\subsection{Translator Module Ablation}
\revv{We evaluate whether the non-linear translator can be replaced by a simpler projection.
Replacing the translator with a linear projection under the same input normalization reduces performance from $59.1$ to $49.3$ mAP$^{50\text{-}95}$ on PerSeg and from $36.3$ to $29.4$ on POD 1-shot.
On held-out regions, the linear projection also yields a higher MSE ($1.24$ vs. $1.09$) and a much weaker teacher-student distance correlation ($0.300$ vs. $0.755$).
This confirms that the translator is not merely a dimensionality adapter, but bridges a non-linear semantic gap between detector features and VLM embeddings.}

\subsection{Dimensionality Reduction}\label{app:dimred}
\begin{figure*}[t]
    \centering
    \includegraphics[width=\textwidth]{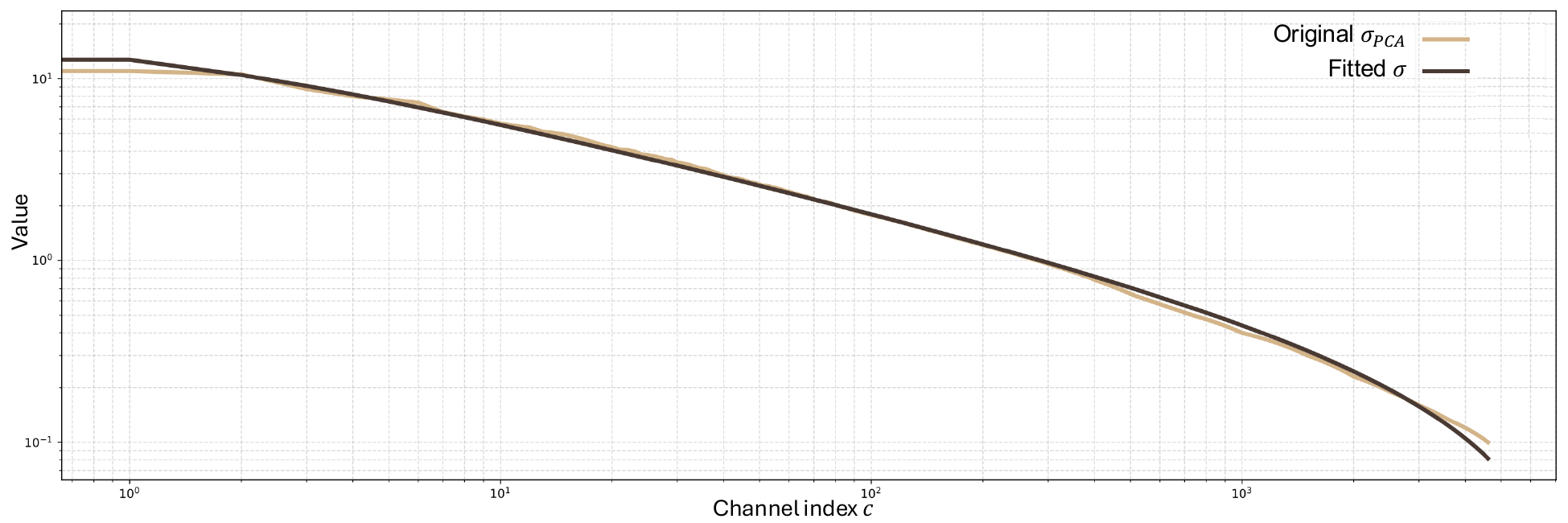}
    \caption{Fitting the PCA curve $\sigma \simeq \frac{a}{(x+1)^b} + c$, $a=18, b=0.47, c=-0.26$.}
    \label{fig:pca-fitting}
\end{figure*}

Fig.~\ref{fig:pca-fitting} shows the variance profile of teacher embeddings, with a fitted decay curve of the form $\sigma \simeq \frac{a}{(x+1)^b} + c$, confirming the typical power-law structure that motivates dimensionality reduction.
The standard deviation of each channel activation in the distillation dataset is estimated and used to rescale the channel $c$ as: $\sigma_c \simeq \frac{18}{(c+1)^{0.47}} - 0.26$ (Eq.~3 in the main document).
\Cref{tab:pca_oi,tab:pca_oi_norm,tab:pca_ref} report extensive results across all datasets and protocols for different PCA settings under three configurations:
\begin{itemize}
\item PCA on $\hat{u}'_i$ in Personal Dataset (Tab.~\ref{tab:pca_ref}), where PCA is applied on teacher embeddings of each personal dataset (ideal case);
\item PCA on $\hat{u}'_i$ in Distillation Dataset (Tab.~\ref{tab:pca_oi}), where PCA is applied on teacher embeddings of distillation dataset;
\item PCA on $u'_i$ in Distillation Dataset (Tab.~\ref{tab:pca_oi_norm}), where the channel normalization of Fig.~\ref{fig:pca-fitting} is applied prior to PCA.
\end{itemize}
Across all experiments, we observe that moderate compression (\eg, retaining $128$–$256$ dimensions) achieves competitive performance, with minimal loss relative to full-dimensional embeddings, while substantially reducing memory and compute costs. As expected, aggressive compression (\eg, $16$–$32$ dimensions) leads to consistent drops across datasets, particularly on POD and iCubWorld.
Overall, this analysis confirms that PCA acts as an effective mechanism to control embedding size while preserving sufficient discriminative information for distillation. The results validate the robustness of our approach to dimensionality reduction, supporting the choice of PCA dimension $d_t=512$.
\revv{To verify that PCA does not collapse the fused representation into a single modality, we analyze the relative contribution of each modality to the retained 512 components.
We compute this contribution by summing squared PCA loadings associated with the dimensions of $h_i$ and $z_{V,i}$.
The analysis assigns 31.7\% of the contribution to $h_i$ and 68.3\% to $z_{V,i}$, confirming that the fused target preserves both high-level semantic and fine-grained visual information.}

\rev{
\subsection{Relational Loss}
To further clarify the role of the relational embedding loss, we report in Tab.~\ref{tab:rel_scaling} an additional ablation that explores lightweight approximations based on Top-$K$ or Rand-$K$ pair sampling, as well as alternative normalization choices.
While sampling can reduce the cost of computing pairwise relations, it consistently underperforms the full formulation, and the gap remains non-negligible on both PerSeg and POD.
In particular, increasing $K$ yields only marginal accuracy gains, suggesting that partial relational views are insufficient to faithfully preserve the global geometry of the teacher space.
Likewise, switching from the proposed $\mathrm{dist}/\sqrt{d}$ scaling to $\ell_2$-normalization provides limited benefits and does not close the performance gap.
Overall, these results support our design choice: leveraging all pairs delivers the most reliable improvements, whereas sampling mainly trades accuracy for modest efficiency gains in our setting.
}

\begin{table}
\centering\small
\caption{\rev{Extra ablation on relational loss configurations.}}
\label{tab:rel_scaling}
\setlength{\tabcolsep}{2.5pt}
\resizebox{0.6\textwidth}{!}{
\begin{tabular}{lccccccc}
\toprule
\multirow{2}{*}{\textbf{Sampling}} & \multirow{2}{*}{\textbf{Norm}} & \multirow{2}{*}{\textbf{K}} & \textbf{PerSeg} & {\textbf{POD}} & \multirow{2}{*}{\textbf{ms/it}$\,\downarrow$} & \multirow{2}{*}{\textbf{VRAM}$\,\downarrow$} \\
& & & {\sc 1 shot} & {\sc 1 shot} & \\
\cmidrule(lr){1-3}\cmidrule(lr){4-4}\cmidrule(lr){5-5}\cmidrule(lr){6-7}
Top-K & dist$/\sqrt{d}$ & 16 & 50.9 \pms{3.4} & 33.6 \pms{2.9} & 324.7 & 19.6\,GB \\ %
Top-K & dist$/\sqrt{d}$ & 32 & 51.1 \pms{3.3} & 32.2 \pms{2.5} & 316.5 & 19.6\,GB \\ %
Rand-K & dist$/\sqrt{d}$ & 32 & 48.3 \pms{3.3} & 32.1 \pms{2.9} & 349.5 & 19.6\,GB \\ %
Top-K & $\ell_2$-norm & 32 & 51.1 \pms{3.3} & 32.4 \pms{2.8} & 318.1 & 19.6\,GB \\
Top-K & -- & 32 & 48.2 \pms{2.9} & 30.4 \pms{3.0} & 342.4 & 19.6\,GB \\
\midrule
\textbf{MOCHA} & dist$/\sqrt{d}$ & all & \textbf{59.1} \pms{3.3} & \textbf{36.3} \pms{3.3} & 347.3 & 19.7\,GB \\
\bottomrule
\end{tabular}%
}%
\end{table}

\renewcommand{\algorithmicensure}{\textbf{Input:}}
\renewcommand{\algorithmiccomment}[1]{~~\# #1}
\begin{algorithm}[t] \small
\caption{Pseudocode for MOCHA's feature distillation.}
\label{alg:distil}
\begin{algorithmic}[1]
\renewcommand{\COMMENT}[1]{\hfill\textcolor{mocha-mid}{\footnotesize\ttfamily\#~#1}}
\REQUIRE Pretrained detection model $\mathrm{m}_{S} = \mathrm{l}_{S} \circ \mathrm{g}_{S}$, visual encoder $\mathrm{g}_{T}$, language model $\mathrm{f}_{T}$ dataset $\mathcal{D}_c$, translation module $\mathrm{t}_{S}$, batch size $K$, number of epochs $E$
\FOR {$e \gets 1\dots E$} 
    \FOR {$k \gets 1\dots \lfloor |\mathcal{D}_c| / K \rfloor$}
    \STATE Sample $\mathcal{K}\sim\mathcal{D}_c$, with $|\mathcal{K}|=K$
        \STATE Initialize batch loss $l \gets 0$
        \FOR {$(X, \mathcal{Y}) \in \mathcal{K}$}
            \STATE $\{F_j\} \gets \mathrm{g}_{S}(X)$ 
            \COMMENT{Compute detection features}
            \STATE $\hat{\mathcal{Y}} \gets \mathrm{l}_{S}(\{F_j\})$
            \COMMENT{Compute detection predictions}
            \STATE $l \gets l + \mathcal{L}_{\mathrm{det}} (\hat{\mathcal{Y}}, \mathcal{Y})$ \COMMENT {Accumulate loss}
            \FOR {$(b_i, c_i) \in \mathcal{Y}$}
                \STATE Use $b_i$ to crop region $X_i$ from $X$
                \STATE $z_{V,i} \gets \mathrm{g}_T (X_i)$ \COMMENT{Vision branch of teacher}
                \STATE $h_{i} \gets \mathrm{f}_T (c_i)$ \COMMENT{Text branch of teacher}
                \STATE Compute $u_i$ from $z_{V,i}$ and $h_{i}$ using Eq.~1
                \STATE Use $b_i$ to crop region $F_{A,j,i}$ from $F_j, \;\; \forall j$
                \STATE $f_{A,i} \gets \mathrm{concat}_j(\text{AvgPool}(F_{A,j,i}))$
                \STATE $f'_{A,i} \gets \mathrm{t}_S(f_{A,i})$ \COMMENT {Translate features}
                \STATE $l \gets l + \mathcal{L}_{\mathrm{dist}} (f'_{A,i}, u_i)/n$ \COMMENT {$n$ = num of boxes}
            \ENDFOR
        \STATE Compute $P_{ff}, P_{uu}$ from $\{f'_{A,i}\}, \{u_i\}$ via Eq.~7
        \STATE $l \gets l + \mathcal{L}_{\mathrm{emb}} (P_{ff}, P_{uu})$
        \ENDFOR
    \STATE $(\mathrm{m}_S, \mathrm{t}_{S}) \gets \text{AdamOptim}(\mathrm{m}_S, \mathrm{t}_{S}, l)$ \COMMENT {Perform gradient descent to update model and translator}
    \ENDFOR
\ENDFOR
\end{algorithmic}
\end{algorithm}

\section{Implementation Details}
All experiments were run on a RHEL8 (RedHat) Unix-based machine with kernel version 6.12.9-1, equipped with 8 NVIDIA L40S GPUs (48GB of VRAM, CUDA 12.8, Driver 570.86.15), 2 AMD EPYC 9224 24-Core processors, 1.5TB of RAM, and Python version 3.11.9. 
Each pretraining experiment uses 4 GPUs, 24 CPU cores, and 64GB of RAM, while for few-shot training and inference, a single GPU and 6 CPU cores are sufficient.
A complete MOCHA distillation (50 epochs) takes about 28 hours, which decreases to around 11 hours when distilling from the AuXFT checkpoints (20 epochs).
Fine-grained information on all packages used in the environment is available as a \texttt{requirements.txt} file in the code repository.

\subsection{Retrieval Evaluation Protocol}
\revv{For CORe50 and iCubWorld, we follow the retrieval-based few-shot personalization protocol of AuXFT.
Given a query instance, the detector first produces candidate proposals.
Each proposal is embedded by the frozen student and translator, then ranked by similarity to the support prototypes.
A query is counted as correct if the target object appears among the top-ranked retrieved proposals.
This protocol is used only for CORe50 and iCubWorld; PerSeg and POD are evaluated with standard end-to-end mAP$^{50\text{-}95}$.}

\subsection{MOCHA Pseudocode}
Algorithm~\ref{alg:distil} reports the steps necessary to distill visual-language knowledge into an efficient vision-only detector, while Algorithms~\ref{alg:fsl_train} and~\ref{alg:fsl_inference} explain how to train MOCHA's few-shot module and perform inference, respectively. Note that lines beginning with the symbol \# denote comments.

\renewcommand{\algorithmicensure}{\textbf{Input:}}
\begin{algorithm} \small
\caption{Pseudocode for MOCHA's FSL training.}
\label{alg:fsl_train}
\begin{algorithmic}[1]
\renewcommand{\COMMENT}[1]{\hfill\textcolor{mocha-mid}{\footnotesize\ttfamily\#~#1}}
\REQUIRE Frozen distilled backbone $\mathrm{g}_S$, frozen translator $\mathrm{t}_S$, prototype classifier $\mathrm{p}(\cdot)$, personal dataset $\mathcal{D}_f$ (train) %
\STATE Initialize empty support set $\mathcal{S} \gets \emptyset$
\FOR[$\mathcal{Y} \subset \mathcal{B} \times \mathcal{C}_c$] {$(X, \mathcal{Y}) \in \mathcal{D}_f$}
    \STATE $\{F_j\} \gets \mathrm{g}_S(X)$ \COMMENT{Extract multiscale features}
    \FOR {$(b_i, c_i) \in \mathcal{Y}$}
        \STATE Use $b_i$ to crop region $F_{A,j,i}$ from $F_j, \;\; \forall j$
        \STATE $f_{A,i} \gets \mathrm{concat}_j(\text{AvgPool}(F_{A,j,i}))$
        \STATE $f'_{A,i} \gets \mathrm{t}_S(f_{A,i})$ \COMMENT{Translated feature}
        \STATE Add pair $(f'_{A,i}, c_i)$ to support set $\mathcal{S}$
    \ENDFOR
\ENDFOR
\STATE Add $\mathcal{S}$ to $\mathrm{p}(\cdot)$'s prototypes store, internal to the local device \COMMENT {Prototype classifier training}
\end{algorithmic}
\end{algorithm}

\renewcommand{\algorithmicensure}{\textbf{Input:}}
\begin{algorithm} \small
\caption{Pseudocode for MOCHA's FSL inference.}
\label{alg:fsl_inference}
\begin{algorithmic}[1]
\renewcommand{\COMMENT}[1]{\hfill\textcolor{mocha-mid}{\footnotesize\ttfamily\#~#1}}
\REQUIRE Frozen distilled model $\mathrm{m}_{S} = \mathrm{l}_{S} \circ \mathrm{g}_{S}$, frozen translator $\mathrm{t}_S$, trained prototype classifier $\mathrm{p}(\cdot)$, image $X$
\STATE $\{F_j\} \gets \mathrm{g}_S(X)$ \COMMENT{Extract multiscale features}
\STATE $\hat{\mathcal{Y}} \gets \mathrm{l}_{S}(\{F_j\})$ \COMMENT {Compute detection predictions for general classes}
\FOR[$\hat{\mathcal{Y}} \subset \mathcal{B} \times \mathcal{C}_c$] {$(\hat{b}_i, \hat{c}_i) \in \hat{\mathcal{Y}}$}
    \STATE Use $\hat{b}_i$ to crop region $F_{A,j,i}$ from $F_j \;\; \forall j$
    \STATE $f_{A,i} \gets \mathrm{concat}_j(\text{AvgPool}(F_{A,j,i}))$
    \STATE $f'_{A,i} \gets \mathrm{t}_S(f_{A,i})$ \COMMENT{Translate to shared space}
    \STATE $\hat{c}_i \gets \mathrm{p}(f'_{A,i})$ \COMMENT{Update class via prototype matching over fine-grained class set}
\ENDFOR
\RETURN Predictions $\hat{\mathcal{Y}}$ \COMMENT {Same boxes, personal classes}
\end{algorithmic}
\end{algorithm}

\begin{table*}[t]
\centering \small
\caption{PCA on $\hat{u}'_i$ in Personal Dataset (mAP/mAcc ± std). Last column (Avg) reports average across the 7 columns. Overall best in bold.}
\label{tab:pca_ref}
\setlength{\tabcolsep}{2.5pt}
\resizebox{\textwidth}{!}{%
\begin{tabular}{lccccccccc}
\toprule
 \multirow{2}{*}{\textbf{Model}} 
         & \multicolumn{1}{c}{\textbf{PerSeg}} 
         & \multicolumn{2}{c}{\textbf{POD}} 
         & \multicolumn{2}{c}{\textbf{CORe50}} 
         & \multicolumn{2}{c}{\textbf{iCubWorld}} 
         &\multirow{2}{*}{\textbf{Avg}}  \\
        & {\sc 1 shot} & {\sc 1 shot} & {\sc 5 shot} & {\sc 1 shot} & {\sc 5 shot} & {\sc 1 shot} & {\sc 5 shot} & \\
\midrule
LLaVa ($u_i$) ($4608$ channels) & 94.0 \pms{1.8} & 66.4 \pms{3.8} & 80.7 \pms{0.0} & 59.4 \pms{4.6} & 80.4 \pms{4.3} & 58.0 \pms{3.2} & 80.6 \pms{2.4} & 74.2 \\
PCA @ $1024$ dims \hfill ($4.5\times$ compression) & 94.2 \pms{1.7} & 67.9 \pms{3.4} & 80.6 \pms{0.0} & 60.1 \pms{4.6} & 80.8 \pms{4.3} & 59.3 \pms{3.1} & 81.2 \pms{2.3} & \textbf{74.9} \\
PCA @ $512$ dims \hfill ($9\times$ compression) & 94.3 \pms{1.7} & 67.9 \pms{3.5} & 80.4 \pms{0.0} & 60.3 \pms{4.6} & 80.6 \pms{4.3} & 59.3 \pms{3.1} & 80.9 \pms{2.3} & 74.8 \\
PCA @ $384$ dims \hfill ($12\times$ compression) & 94.4 \pms{1.7} & 67.5 \pms{3.4} & 80.0 \pms{0.0} & 60.4 \pms{4.6} & 80.6 \pms{4.3} & 59.3 \pms{3.0} & 80.8 \pms{2.2} & 74.7 \\
PCA @ $256$ dims \hfill ($18\times$ compression) & 94.2 \pms{1.6} & 67.6 \pms{3.3} & 79.6 \pms{0.0} & 60.6 \pms{4.6} & 80.5 \pms{4.5} & 59.2 \pms{3.1} & 80.2 \pms{2.2} & 74.6 \\
PCA @ $128$ dims \hfill ($36\times$ compression) & 92.9 \pms{1.5} & 66.4 \pms{3.1} & 79.4 \pms{0.0} & 60.5 \pms{4.6} & 79.8 \pms{4.3} & 57.5 \pms{3.1} & 77.6 \pms{2.4} & 73.4 \\
PCA @ $64$ dims \hfill ($72\times$ compression) & 88.0 \pms{2.9} & 63.9 \pms{3.4} & 79.9 \pms{0.0} & 59.0 \pms{4.5} & 77.1 \pms{4.6} & 52.7 \pms{3.3} & 71.2 \pms{2.4} & 70.3 \\
PCA @ $32$ dims \hfill ($144\times$ compression) & 81.5 \pms{2.8} & 60.0 \pms{3.5} & 70.6 \pms{0.0} & 53.5 \pms{4.7} & 71.4 \pms{5.0} & 45.9 \pms{3.1} & 62.1 \pms{2.7} & 63.6 \\
\bottomrule
\end{tabular}%
}%
\end{table*}

\begin{table*}[t]
\centering \small
\caption{PCA on $\hat{u}'_i$ in Distillation Dataset (mAP/mAcc ± std). Last column (Avg) reports average across the 7 columns. Overall best in bold.}
\label{tab:pca_oi}
\setlength{\tabcolsep}{2.5pt}
\resizebox{\textwidth}{!}{%
\begin{tabular}{lccccccccc}
\toprule
 \multirow{2}{*}{\textbf{Model}} 
         & \multicolumn{1}{c}{\textbf{PerSeg}} 
         & \multicolumn{2}{c}{\textbf{POD}} 
         & \multicolumn{2}{c}{\textbf{CORe50}} 
         & \multicolumn{2}{c}{\textbf{iCubWorld}} 
         &\multirow{2}{*}{\textbf{Avg}}  \\
        & {\sc 1 shot} & {\sc 1 shot} & {\sc 5 shot} & {\sc 1 shot} & {\sc 5 shot} & {\sc 1 shot} & {\sc 5 shot} & \\
\midrule
LLaVa ($u_i$) ($4608$ channels) & 94.0 \pms{1.8} & 66.4 \pms{3.8} & 80.7 \pms{0.0} & 59.4 \pms{4.6} & 80.4 \pms{4.3} & 58.0 \pms{3.2} & 80.6 \pms{2.4} & 74.2 \\
PCA @ $1024$ dims \hfill ($4.5\times$ compression) & 94.4 \pms{1.7} & 67.9 \pms{3.5} & 80.3 \pms{0.0} & 59.7 \pms{4.5} & 80.3 \pms{4.3} & 59.2 \pms{3.3} & 81.2 \pms{2.2} & \textbf{74.7} \\
PCA @ $512$ dims \hfill ($9\times$ compression) & 94.2 \pms{1.6} & 67.8 \pms{3.5} & 80.1 \pms{0.0} & 59.8 \pms{4.5} & 80.1 \pms{4.3} & 58.9 \pms{3.3} & 80.8 \pms{2.2} & 74.5 \\
PCA @ $384$ dims \hfill ($12\times$ compression) & 93.7 \pms{1.7} & 67.7 \pms{3.4} & 80.4 \pms{0.0} & 59.8 \pms{4.6} & 79.9 \pms{4.4} & 58.6 \pms{3.3} & 80.3 \pms{2.3} & 74.3 \\
PCA @ $256$ dims \hfill ($18\times$ compression) & 93.0 \pms{1.6} & 67.4 \pms{3.5} & 79.2 \pms{0.0} & 59.3 \pms{4.6} & 79.6 \pms{4.4} & 57.6 \pms{3.3} & 79.3 \pms{2.2} & 73.6 \\
PCA @ $128$ dims \hfill ($36\times$ compression) & 90.4 \pms{2.0} & 65.9 \pms{3.1} & 78.6 \pms{0.0} & 56.7 \pms{4.7} & 77.2 \pms{4.5} & 54.3 \pms{3.3} & 75.3 \pms{2.3} & 71.2 \\
PCA @ $64$ dims \hfill ($72\times$ compression) & 85.8 \pms{3.1} & 62.5 \pms{3.3} & 74.8 \pms{0.0} & 51.9 \pms{4.6} & 72.2 \pms{4.7} & 48.4 \pms{3.2} & 67.9 \pms{2.5} & 66.2 \\
PCA @ $32$ dims \hfill ($144\times$ compression) & 79.3 \pms{3.2} & 57.0 \pms{3.6} & 72.1 \pms{0.0} & 44.8 \pms{4.7} & 63.4 \pms{5.1} & 43.7 \pms{3.2} & 60.7 \pms{2.6} & 60.1 \\
\bottomrule
\end{tabular}%
}%
\end{table*}

\begin{table*}[t]
\centering \small
\caption{PCA on $u'_i$ in Distillation Dataset (mAP/mAcc ± std). Last column (Avg) reports average across the 7 columns. Overall best in bold.}
\label{tab:pca_oi_norm}
\setlength{\tabcolsep}{4pt}
\resizebox{\textwidth}{!}{%
\begin{tabular}{lccccccccc}
\toprule
 \multirow{2}{*}{\textbf{Model}} 
         & \multicolumn{1}{c}{\textbf{PerSeg}} 
         & \multicolumn{2}{c}{\textbf{POD}} 
         & \multicolumn{2}{c}{\textbf{CORe50}} 
         & \multicolumn{2}{c}{\textbf{iCubWorld}} 
         &\multirow{2}{*}{\textbf{Avg}}  \\
        & {\sc 1 shot} & {\sc 1 shot} & {\sc 5 shot} & {\sc 1 shot} & {\sc 5 shot} & {\sc 1 shot} & {\sc 5 shot} & \\
\midrule
LLaVa ($u_i$) ($4608$ channels) & 95.1 \pms{1.7} & 70.0 \pms{3.4} & 80.6 \pms{0.0} & 57.0 \pms{4.0} & 79.7 \pms{3.7} & 56.4 \pms{3.6} & 81.2 \pms{2.5} & 74.3 \\
PCA @ $1024$ dims \hfill ($4.5\times$ compression) & 96.6 \pms{1.3} & 68.5 \pms{3.7} & 81.7 \pms{0.0} & 58.9 \pms{4.3} & 80.6 \pms{3.9} & 61.2 \pms{3.6} & 83.4 \pms{2.2} & 75.8 \\
PCA @ $512$ dims \hfill ($9\times$ compression) & 96.9 \pms{1.1} & 68.9 \pms{3.7} & 81.9 \pms{0.0} & 59.6 \pms{4.3} & 80.6 \pms{4.0} & 63.7 \pms{3.4} & 85.5 \pms{2.1} & 76.7 \\
PCA @ $384$ dims \hfill ($12\times$ compression) & 96.4 \pms{1.2} & 68.5 \pms{3.7} & 83.3 \pms{0.0} & 60.4 \pms{4.4} & 80.6 \pms{4.2} & 64.0 \pms{3.2} & 86.0 \pms{2.0} & \textbf{77.0} \\
PCA @ $256$ dims \hfill ($18\times$ compression) & 95.2 \pms{1.3} & 68.6 \pms{3.6} & 83.2 \pms{0.0} & 60.9 \pms{4.4} & 81.0 \pms{4.2} & 64.1 \pms{3.3} & 86.0 \pms{1.9} & \textbf{77.0} \\
PCA @ $128$ dims \hfill ($36\times$ compression) & 92.2 \pms{1.9} & 67.7 \pms{3.7} & 79.3 \pms{0.0} & 59.9 \pms{4.4} & 79.6 \pms{4.3} & 62.6 \pms{3.2} & 84.2 \pms{1.9} & 75.1 \\
PCA @ $64$ dims \hfill ($72\times$ compression) & 87.9 \pms{2.2} & 65.7 \pms{3.6} & 75.7 \pms{0.0} & 55.3 \pms{4.5} & 74.5 \pms{4.7} & 55.0 \pms{3.1} & 76.4 \pms{2.4} & 70.1 \\
PCA @ $32$ dims \hfill ($144\times$ compression) & 80.2 \pms{3.2} & 60.6 \pms{3.8} & 74.6 \pms{0.0} & 46.4 \pms{4.6} & 64.7 \pms{5.5} & 48.0 \pms{3.2} & 68.0 \pms{2.8} & 63.2 \\
\bottomrule
\end{tabular}%
}%
\end{table*}

\clearpage
\newpage

\begin{figure*}[t]
    \centering
    \begin{subfigure}{\linewidth}
        \rotatebox{90}{~YOLOv8n}
        \begin{subfigure}{.3\textwidth}
            \includegraphics[width=0.49\textwidth]{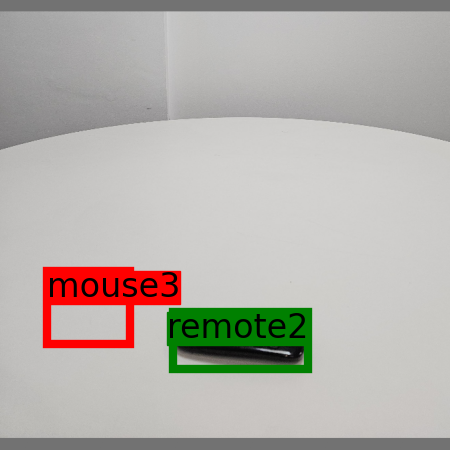}
            \includegraphics[width=0.49\textwidth]{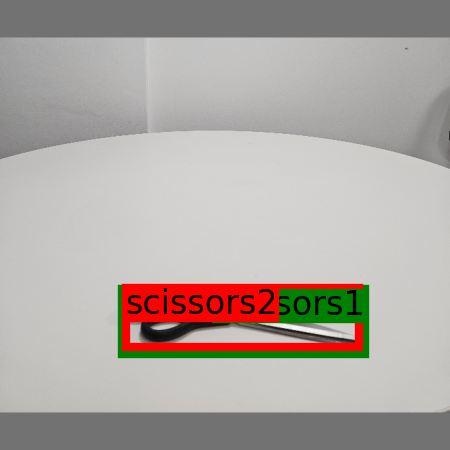}
        \end{subfigure}
        \begin{subfigure}{.3\textwidth}
            \includegraphics[width=0.49\textwidth]{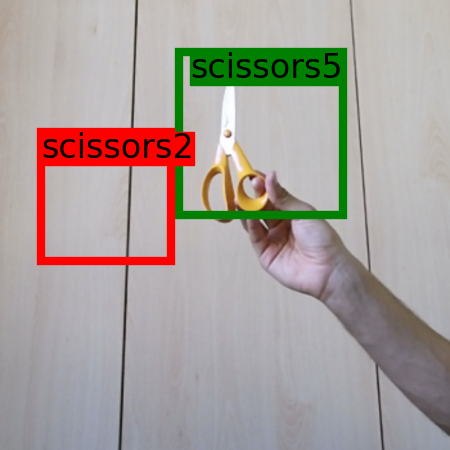}
            \includegraphics[width=0.49\textwidth]{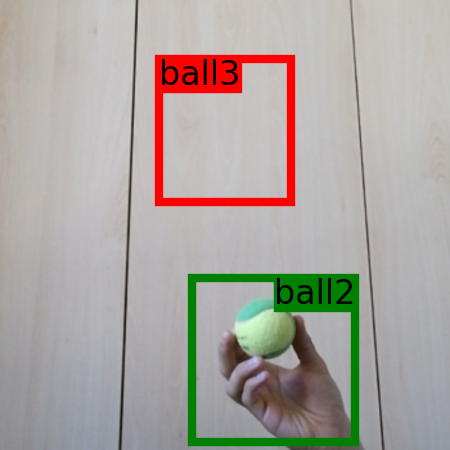}
        \end{subfigure}
        \begin{subfigure}{.3\textwidth}
            \includegraphics[width=0.49\textwidth]{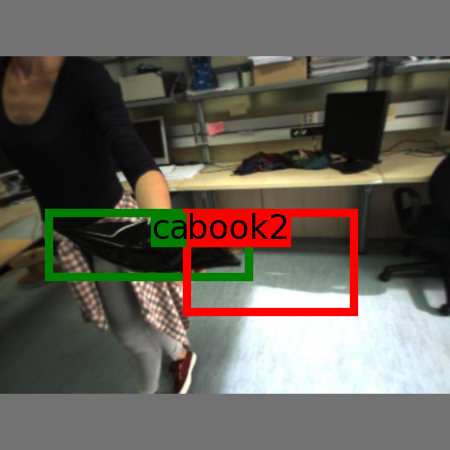}
            \includegraphics[width=0.49\textwidth]{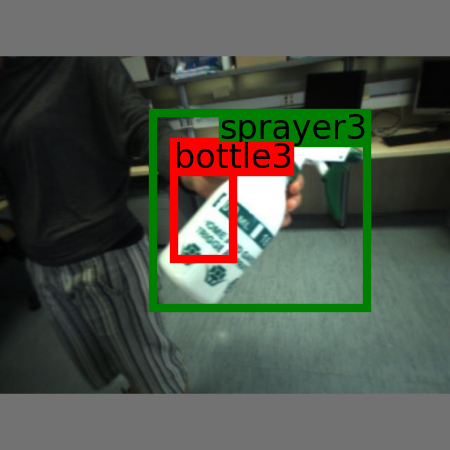}
        \end{subfigure}
    \end{subfigure}
    \begin{subfigure}{\linewidth}
        \rotatebox{90}{~~~AuXFT}
        \begin{subfigure}{.3\textwidth}
            \includegraphics[width=0.49\textwidth]{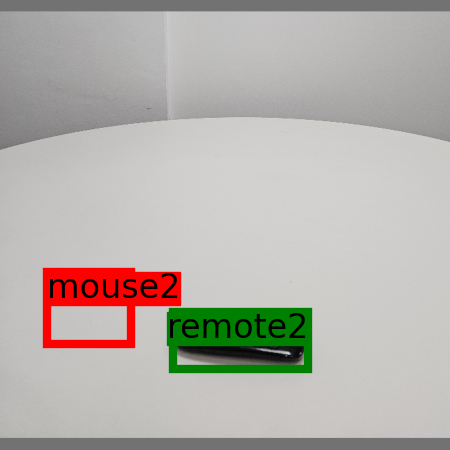}
            \includegraphics[width=0.49\textwidth]{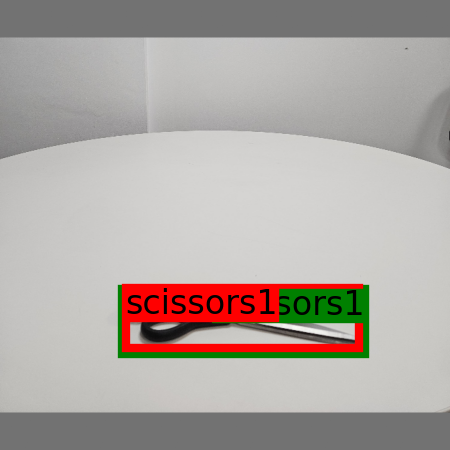}
        \end{subfigure}
        \begin{subfigure}{.3\textwidth}
            \includegraphics[width=0.49\textwidth]{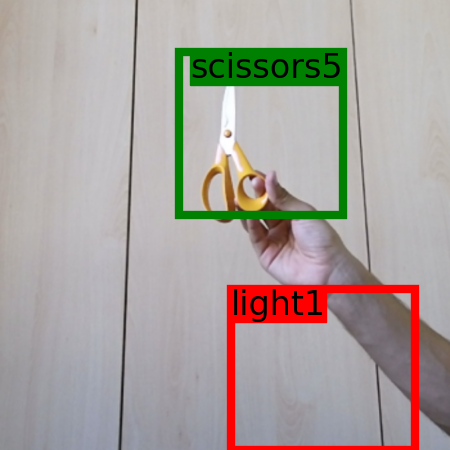}
            \includegraphics[width=0.49\textwidth]{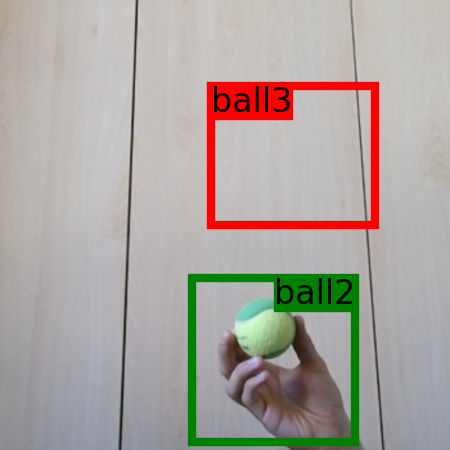}
        \end{subfigure}
        \begin{subfigure}{.3\textwidth}
            \includegraphics[width=0.49\textwidth]{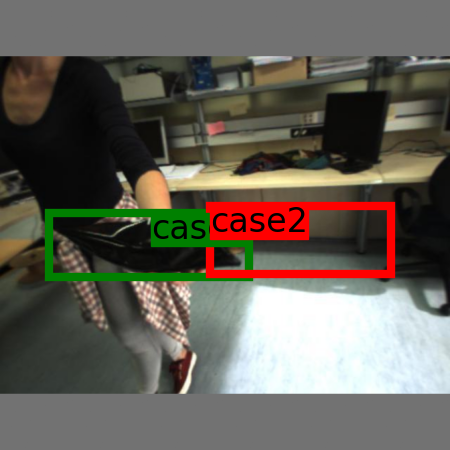}
            \includegraphics[width=0.49\textwidth]{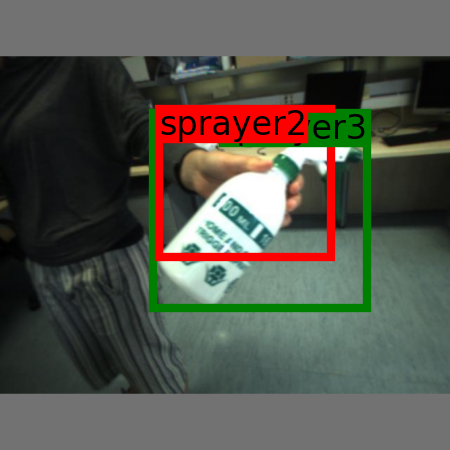}
        \end{subfigure}
    \end{subfigure}
\begin{subfigure}{\linewidth}
        \rotatebox{90}{~~~~~MOCHA*} 
        \begin{subfigure}{.3\textwidth}
            \includegraphics[width=0.49\textwidth]{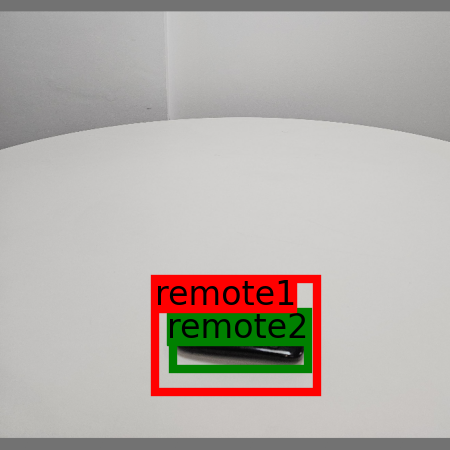}
            \includegraphics[width=0.49\textwidth]{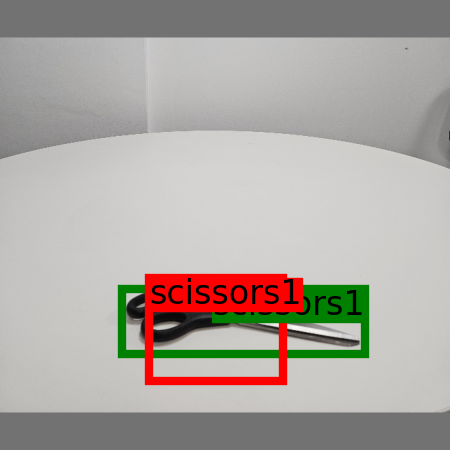}
            \caption*{POD}
        \end{subfigure}
        \begin{subfigure}{.3\textwidth}
            \includegraphics[width=0.49\textwidth]{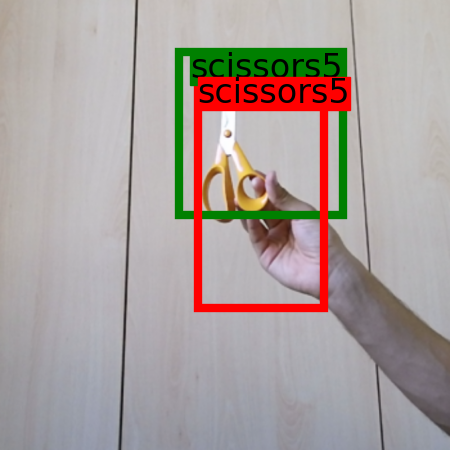}
            \includegraphics[width=0.49\textwidth]{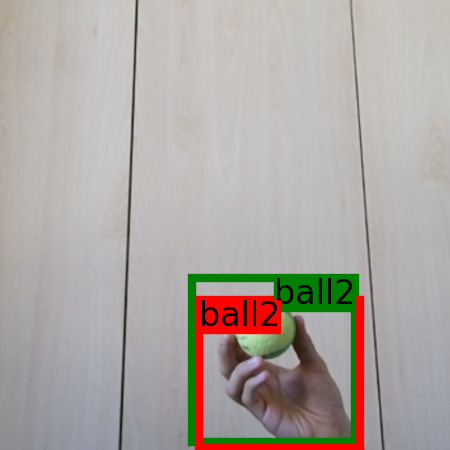}
            \caption*{CoRE50}
        \end{subfigure}
        \begin{subfigure}{.3\textwidth}
            \includegraphics[width=0.49\textwidth]{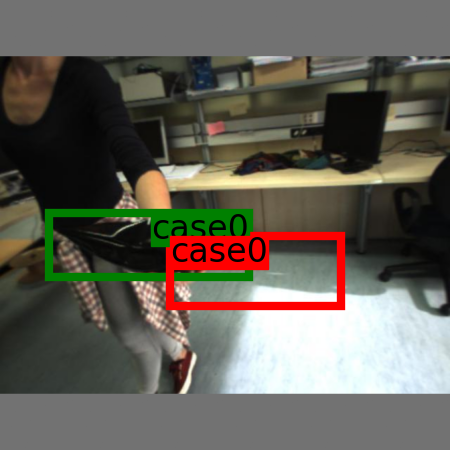}
            \includegraphics[width=0.49\textwidth]{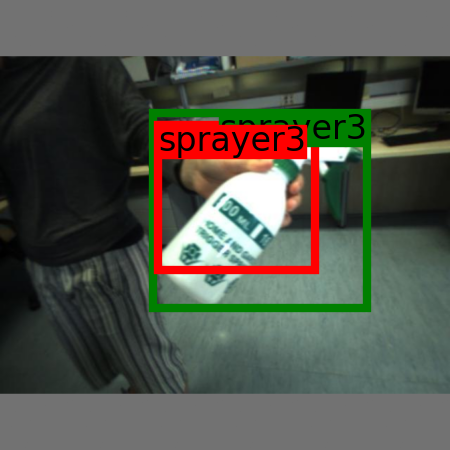}
            \caption*{iCubWorld}
        \end{subfigure}
    \end{subfigure}
    \caption{Qualitative results in the 1-shot setting. Ground truth in green, prediction with the highest confidence in red (class names shown refer to personal class labels in each dataset). YOLOv8n refers to the baseline. *: refers to MOCHA (AuXFT). %
    }
    \label{fig:quali-1}
\end{figure*}

\begin{figure*}[t]
    \centering
    \begin{subfigure}{\textwidth}
        \rotatebox{90}{~YOLOv8n}
        \begin{subfigure}{.3\textwidth}
            \includegraphics[width=0.49\textwidth]{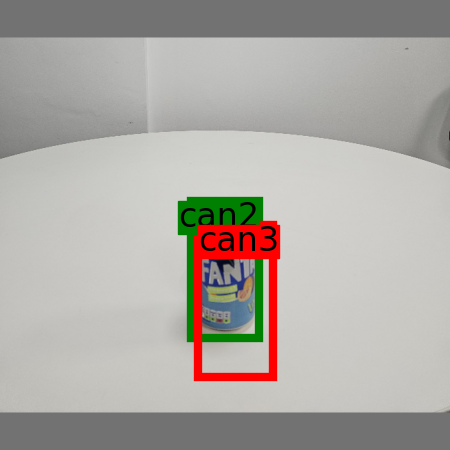}
            \includegraphics[width=0.49\textwidth]{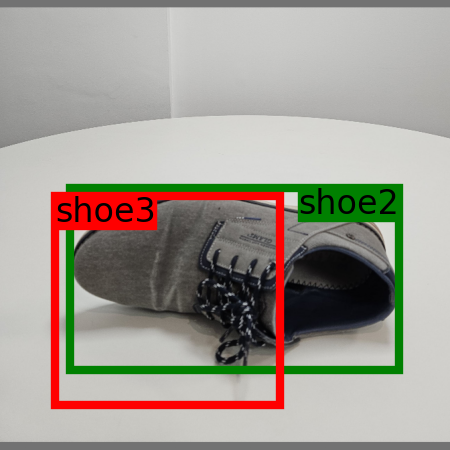}
        \end{subfigure}
        \begin{subfigure}{.3\textwidth}
            \includegraphics[width=0.49\textwidth]{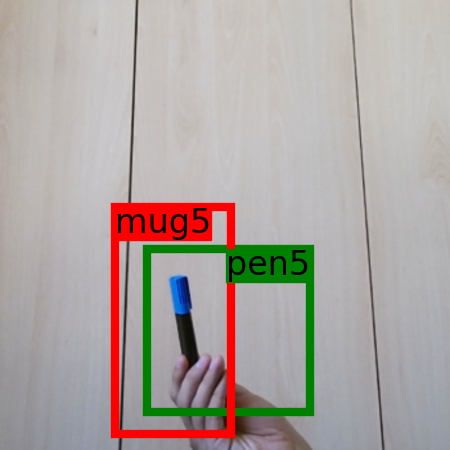}
            \includegraphics[width=0.49\textwidth]{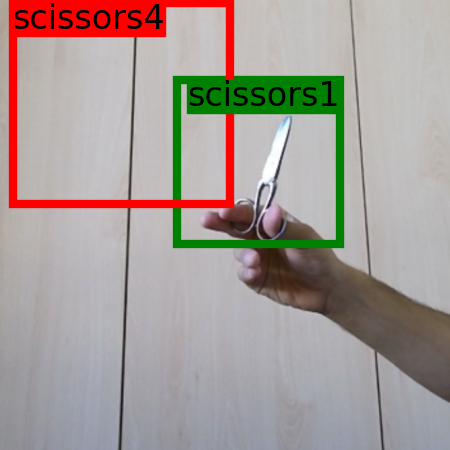}
        \end{subfigure}
        \begin{subfigure}{.3\textwidth}
            \includegraphics[width=0.49\textwidth]{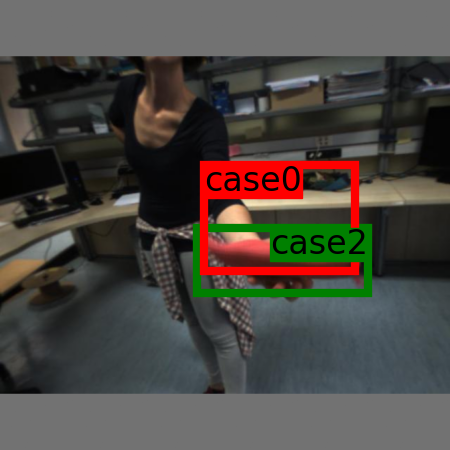}
            \includegraphics[width=0.49\textwidth]{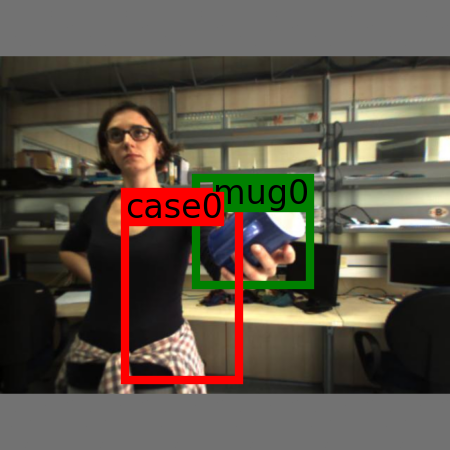}
        \end{subfigure}
    \end{subfigure}
        \begin{subfigure}{\linewidth}
        \rotatebox{90}{~~~AuXFT}
        \begin{subfigure}{.3\textwidth}
            \includegraphics[width=0.49\textwidth]{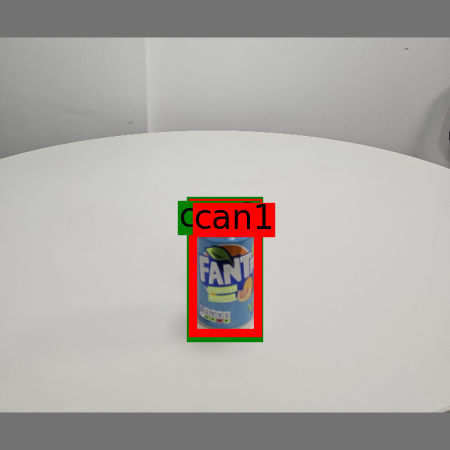}
            \includegraphics[width=0.49\textwidth]{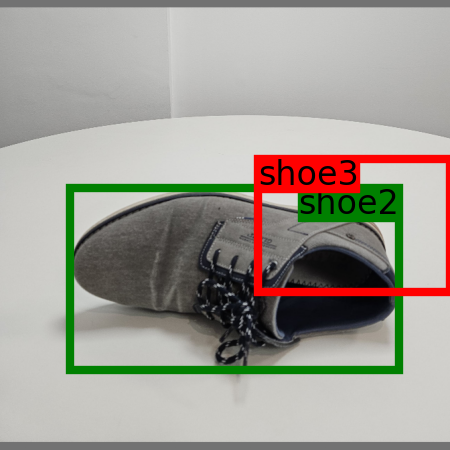}
        \end{subfigure}
        \begin{subfigure}{.3\textwidth}
            \includegraphics[width=0.49\textwidth]{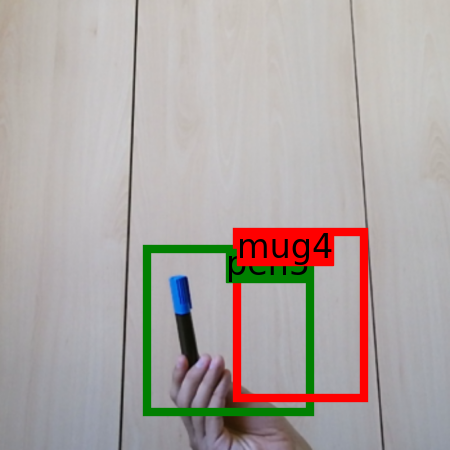}
            \includegraphics[width=0.49\textwidth]{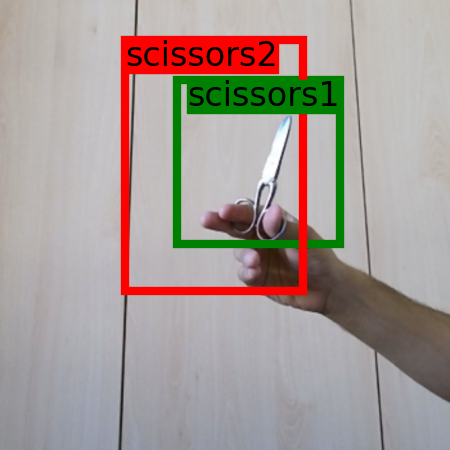}
        \end{subfigure}
        \begin{subfigure}{.3\textwidth}
            \includegraphics[width=0.49\textwidth]{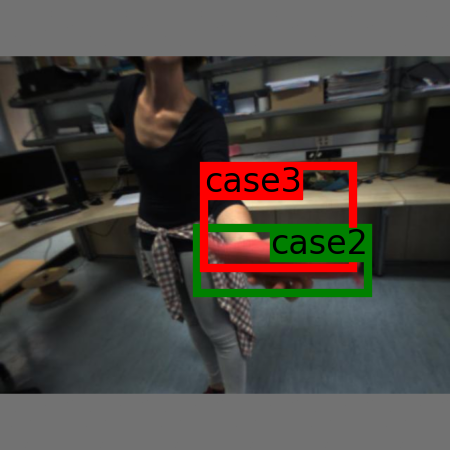}
            \includegraphics[width=0.49\textwidth]{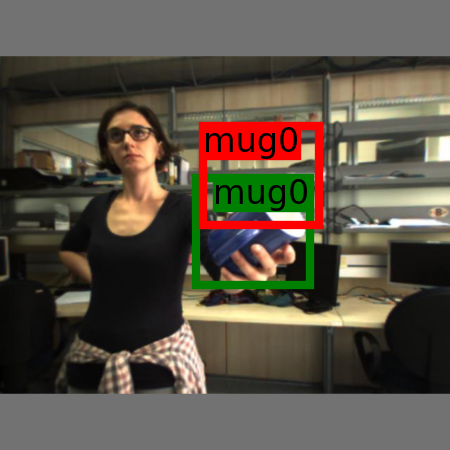}
        \end{subfigure}
    \end{subfigure}
    \begin{subfigure}{\linewidth}
        \rotatebox{90}{~~~~~MOCHA*} 
        \begin{subfigure}{.3\textwidth}
            \includegraphics[width=0.49\textwidth]{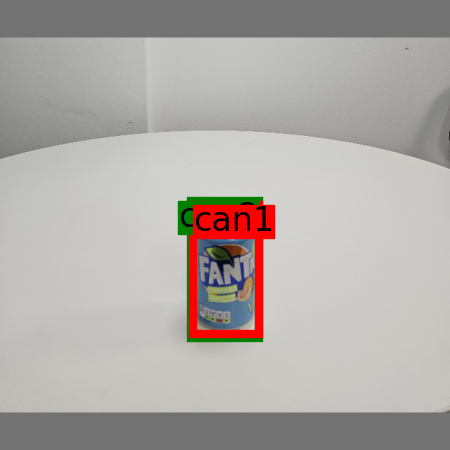}
            \includegraphics[width=0.49\textwidth]{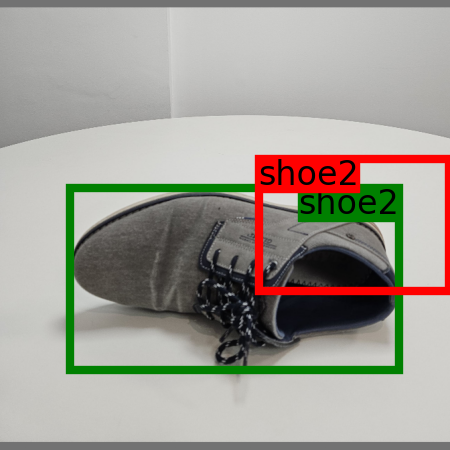}
            \caption*{POD}
        \end{subfigure}
        \begin{subfigure}{.3\textwidth}
            \includegraphics[width=0.49\textwidth]{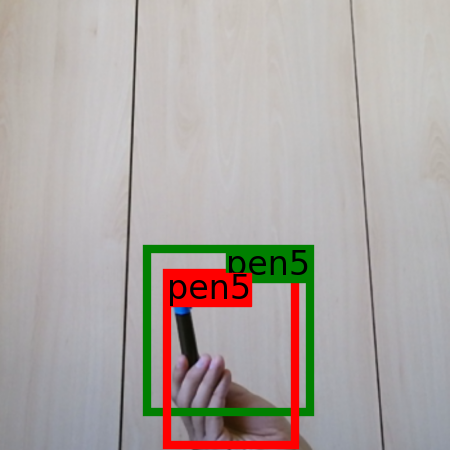}
            \includegraphics[width=0.49\textwidth]{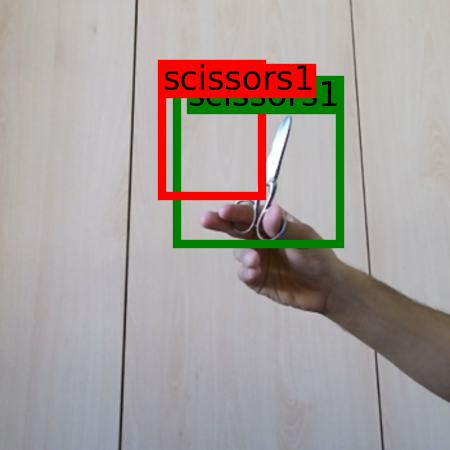}
            \caption*{CoRE50}
        \end{subfigure}
        \begin{subfigure}{.3\textwidth}
            \includegraphics[width=0.49\textwidth]{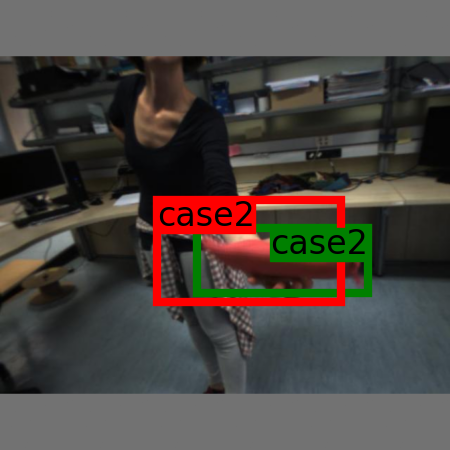}
            \includegraphics[width=0.49\textwidth]{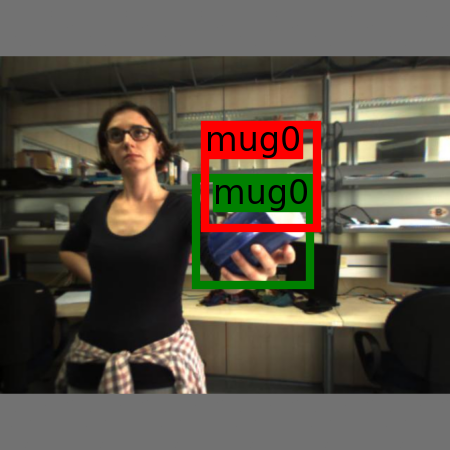}
            \caption*{iCubWorld}
        \end{subfigure}
    \end{subfigure}
    \caption{Qualitative results in the 5-shot setting. Ground truth in green, prediction with the highest confidence in red (class names shown refer to personal class labels in each dataset). YOLOv8n refers to the baseline. *: refers to MOCHA (AuXFT). %
    }
    \label{fig:quali-5}
\end{figure*}

\clearpage
\newpage
\section{Qualitative Results}
\Cref{fig:quali-1,fig:quali-5} present qualitative comparisons across three models---YOLO (baseline), AuXFT, and MOCHA (AuXFT)---in both the 1-shot and 5-shot personalization settings. Results are shown on three representative datasets: POD, CoRE50, and iCubWorld. Ground truth annotations are depicted in green, while the prediction with the highest confidence is shown in red.
In the 1-shot setting (\cref{fig:quali-1}), MOCHA already improves over both YOLO and AuXFT, correctly localizing and classifying objects that other models miss or mislabel (\eg, \textit{remote2} in POD, \textit{case0} in iCubWorld). With 5-shot supervision (\cref{fig:quali-5}), MOCHA further enhances precision, handling subtle distinctions (\eg, \textit{pen5} vs. \textit{mug5} in CoRE50), and resolving multiple similar instances in cluttered scenes.
These examples qualitatively validate MOCHA's effectiveness in aligning student features to the teacher's multimodal embedding space, enabling more reliable object personalization with minimal data.

\section{Limitations}
\ec[]{
While MOCHA brings notable advances, several limitations remain that open directions for future research.
First, MOCHA is inherently tailored to object detection, which may restrict its applicability to other tasks.
Second, although inference incurs minimal overhead, the dependence on large foundation models during training introduces considerable computational cost, which may be impractical in certain scenarios.
Finally, the prototypical network component becomes inefficient at scale, leading to potential performance degradation when handling a large number of personalized object instances.
}

\bibliographystyle{splncs04}
\bibliography{main}

\newpage
\end{document}